\documentclass[10pt,twocolumn,letterpaper]{article}

\usepackage[pagenumbers]{cvpr}

\definecolor{cvprblue}{rgb}{0.21,0.49,0.74}
\usepackage[pagebackref,breaklinks,colorlinks,allcolors=cvprblue]{hyperref}

\usepackage[utf8]{inputenc} 
\usepackage[T1]{fontenc}    
\usepackage{hyperref}       
\usepackage{url}            
\usepackage{booktabs}       
\usepackage{amsfonts}       
\usepackage{nicefrac}       
\usepackage{microtype}      
\usepackage{xcolor}         
\usepackage{graphicx}
\usepackage{caption} 
\usepackage{enumitem}
\usepackage{amsmath} 
\usepackage{amssymb}     
\usepackage[table]{xcolor} 
\usepackage{multirow}
\usepackage{wrapfig}
\usepackage[accsupp]{axessibility}

\def\paperID{} 
\def\confName{CVPR}
\def\confYear{2026}

\title{Enhancing Continual Learning of Vision-Language Models\\via Dynamic Prefix Weighting}

\author{
 Hyeonseo Jang \qquad
 Hyuk Kwon \qquad
 Kibok Lee \\
Yonsei University \\
 {\tt\small \{jhyeonseo715, kh12043, kibok\}@yonsei.ac.kr}
}

\begin{document}
\maketitle
\begin{abstract}
We investigate recently introduced domain-class incremental learning scenarios for vision-language models (VLMs). Recent works address this challenge using parameter-efficient methods, such as prefix-tuning or adapters, which facilitate model adaptation to downstream tasks by incorporating task-specific information into input tokens through additive vectors. However, previous approaches often normalize the weights of these vectors, disregarding the fact that different input tokens require different degrees of adjustment. To overcome this issue, we propose Dynamic Prefix Weighting (DPW), a framework that dynamically assigns weights to prefixes, complemented by adapters. DPW consists of 1)~a gating module that adjusts the weights of each prefix based on the importance of the corresponding input token, and 2)~a weighting mechanism that derives adapter output weights as a residual of prefix-tuning weights, ensuring that adapters are utilized only when necessary. Experimental results demonstrate that our method achieves state-of-the-art performance in domain-class incremental learning scenarios for VLMs.
The code is available at: \url{https://github.com/YonseiML/dpw}.
\end{abstract}    
\section{Introduction}
\label{sec:Introduction}

Continual learning (CL) enables models to adapt to sequential data streams while mitigating catastrophic forgetting of previously acquired knowledge, offering an alternative to retraining from scratch in ever-changing environments~\cite{DeLange2021, Rebuffi2017, Dhar2019}.
Recent advances in CL have extended its applicability to multi-modal learning scenarios~\cite{Yu2024_survey}, including vision-language models (VLMs) such as CLIP~\cite{Radford2021}.
Despite their impressive zero-shot capabilities, VLMs often yield suboptimal performance in real-world applications, underscoring the need for CL to improve downstream task adaptation while preserving the foundational knowledge established during pretraining~\cite{Zheng2023}.
To address this, previous studies have introduced benchmarks evaluating both the zero-shot and CL performance of VLMs on downstream tasks~\cite{Zheng2023, Li2024}.
However, the large-scale nature of VLMs poses significant challenges in fine-tuning the entire model due to the computational cost~\cite{Khattak2023a, Yu2024}.

\begin{figure*}[t!]
\begin{center}
\resizebox{0.9\linewidth}{!}{
  \includegraphics[width=\textwidth]{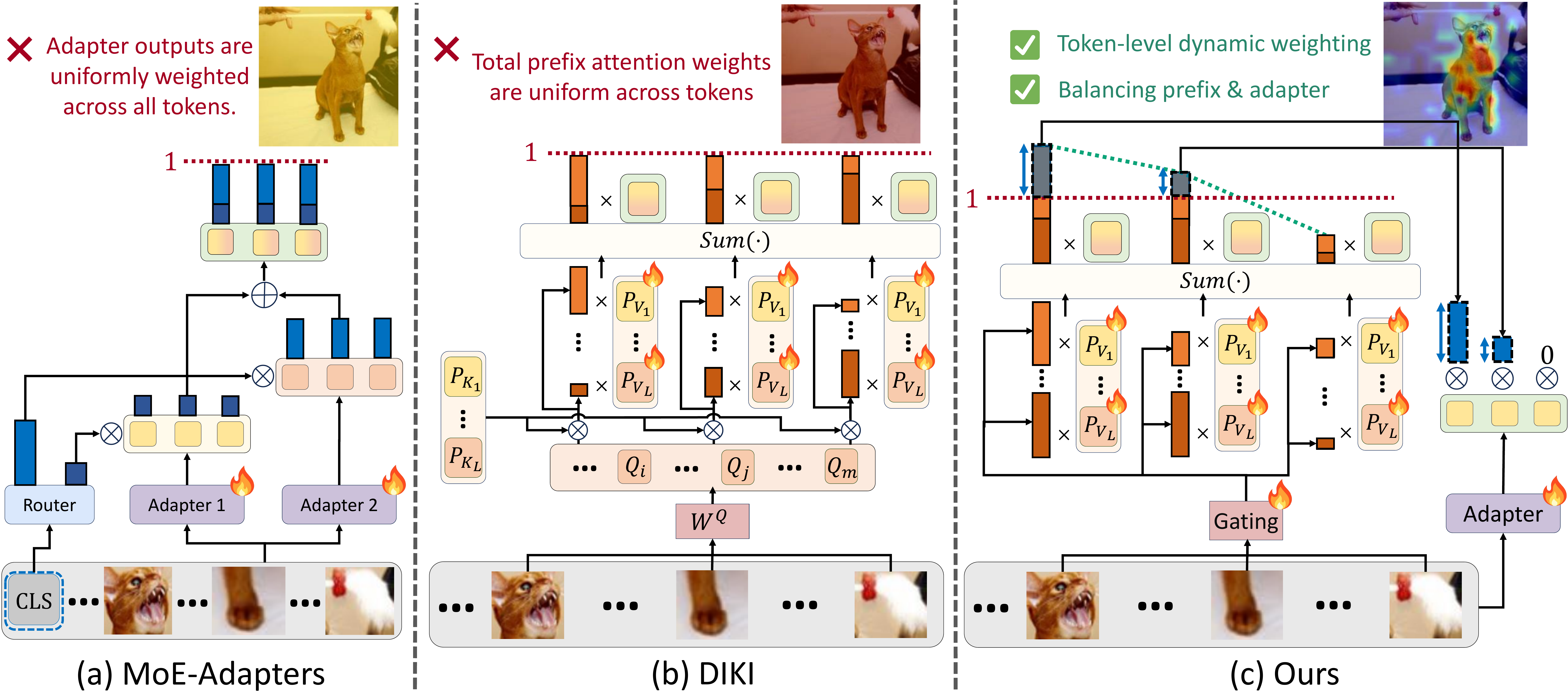}
}
\vspace{-8pt}
\caption{Comparison of the weighting mechanisms of various popular PEFT methods. Light-green boxes denote the output vectors added to the input tokens.
(a): MoE-Adapters~\cite{Yu2024} relies solely on the CLS token for routing and assigns equal weights to the outputs of each adapter, with their sum always fixed at one.
(b): DIKI~\cite{Tang2024} always normalizes the total weight of the prefixes added to each input, forcing them to sum up to one.
(c): Our method dynamically adjusts prefixes and adapters weights based on the importance of the input tokens.}
\label{fig:motivation}
\vspace{-22pt}
\end{center}
\end{figure*}

Recent studies have explored parameter-efficient fine-tuning (PEFT) methods to mitigate this challenge. 
These approaches typically freeze the backbone model and learn task-specific knowledge by introducing lightweight adapters~\cite{Yu2024, C-CLIP2025} or employing prefix-tuning techniques with learnable prompt vectors~\cite{Lu2024, Li2024, Tang2024}. 
Such methods enable effective adaptation to new tasks while requiring only a small number of additional parameters.
In the context of CL, recent PEFT advances primarily focus on adaptively weighting the influence of newly introduced parameters based on input samples, thereby mitigating catastrophic forgetting~\cite{Tang2024, Yu2024}.
However, as illustrated in Fig.~\ref{fig:motivation}, existing weighting mechanisms generally operate at the sample level,
treating all tokens within a sample uniformly and assigning them the same amount of task-specific information.
As prior studies~\cite{Fu2022, Zhou2024} have shown that dynamically adjusting the amount of injected information according to the task relevance of each token can substantially improve performance, the lack of token-level weighting thus represents a fundamental limitation.

We further observe that token-level weighting is hindered not only by the weighting strategies used in existing methods, but also by inherent properties of the pretrained backbone.
For instance, within the prefix-tuning framework, existing methods typically rely on attention mechanisms to assign prefix weights, which can blur the token-wise distinctions required for fine-grained weighting.
Because attention mechanisms tend to capture global contextual relationships within a sample~\cite{Pan2024, Wang2024}, they often project tokens with distinct characteristics closer together in the feature space. 
This behavior obscures the distinctions between task-relevant and irrelevant tokens, making fine-grained weighting difficult.
Moreover, this limitation extends beyond individual samples to the task level, hindering the disentanglement of token embeddings across tasks. 
As a result, knowledge transfer in CL settings becomes less effective, and task interference increasingly degrades overall performance~\cite{Hiratani2024}.

In this paper, we investigate how to incorporate an appropriate amount of task-specific information into each input token in CL.
To this end, we propose dynamic prefix weighting (\textbf{DPW}), a framework that assigns refined weights to both prefixes and adapters based on the task relevance of the corresponding input token.
First, we develop a novel gating module that replaces the conventional attention mechanism used in prefix-tuning.
Our gating module consists of reparametrized prefix attention (\textbf{RePA}) and conditional activation (\textbf{CondAct}), which replace the query-key dot product and softmax activation, respectively.
Unlike conventional prefix-tuning methods that rely on query-key projection, our RePA computes task relevance scores directly from input tokens using a learnable affine transformation.
Subsequently, CondAct replaces the softmax activation with conditional normalization and filtering, ensuring that the total prefix weight is dynamically adjusted within an appropriate range.
Second, we introduce a residual weighting mechanism (\textbf{RWM}) that assigns weights to adapters based on residual prefix weights generated during conditional normalization.
RWM selectively incorporates adapter outputs for input tokens with higher prefix scores, enabling adapters to focus on more informative tokens according to refined prefix scores.
As illustrated in Fig.~\ref{fig:motivation}(c), DPW provides fine-grained weighting for both prefixes and adapters, enabling precise token-level modulation and improving knowledge retention in CL settings.
  
\section{Related Work}

\textbf{Continual Learning (CL).} Existing CL approaches can be categorized into three types: regularization-based, memory-based, and architecture-based methods. \textit{Regularization-based methods}~\cite{Aljundi2018, Aljundi2019a, Kirkpatrick2017, Zenke2017, Lee2017} mitigate catastrophic forgetting by introducing penalty terms into the loss function, thereby preserving the prior knowledge acquired from earlier tasks. \textit{Memory-based approaches}~\cite{Chaudhry2018, Liu2020, Luo2023, Yan2022, Yan2021, Bang2021} store historical data or synthetic samples generated by models, allowing selective replay to retain past knowledge. \textit{Architecture-based methods}~\cite{Douillard2022, Li2019, Mallya2018, Liang2024, Smith2023, Wang2022a, Wang2022b, Wang2022c} expand model parameters to incorporate new tasks while retaining knowledge from previous ones.
PEFT based approaches, such as adapter and prompt-tuning, are typically categorized as architecture-based methods, as they introduce a small set of task-specific parameters while keeping the backbone frozen. Our method adopts a similar strategy, enabling efficient adaptation to downstream tasks with few additional parameters.

\noindent\textbf{Continual Learning of VLMs.} CL of VLMs seeks to enable these models to sequentially adapt to diverse domains and classes while preserving their zero-shot capabilities~\cite{Zheng2023, Li2024}. One of the pioneering approaches is ZSCL~\cite{Zheng2023}, which addresses this challenge by leveraging a reference dataset and employing knowledge distillation from a frozen CLIP. 
Subsequent work has explored advanced distillation strategies, such as incorporating CLIP models trained on previous tasks into the distillation process with sample-wise adaptive balancing between two teacher models~\cite{Chu2024}, or utilizing diffusion models to generate reference datasets for distillation~\cite{GIFT}. 
However, these approaches require fine-tuning the entire CLIP model and rely on an additional reference dataset, which reduces training efficiency.
An alternative line of work leverages PEFT techniques, such as adapters~\cite{Yu2024, C-CLIP2025, Gao2025}, for more efficient adaptation.
For example, MoE-Adapters~\cite{Yu2024} introduce two adapters per task and use a router to assign weights within a mixture-of-experts framework, combining their outputs with the input tokens.  C-CLIP~\cite{C-CLIP2025} learns LoRA with a distillation loss and integrates it into the backbone after training each task. 
Another PEFT branch focuses on prompt-based methods~\cite{Li2024, Tang2024, Lu2024}, which learn task-specific information through a learnable vector called a prompt or a prefix. 
CoLeCLIP~\cite{Li2024} extends to new tasks using a dedicated vocabulary and task-specific prompts.
DIKI~\cite{Tang2024} mitigates prefix interference during CL by using cross-attention to separately assign weights to prefixes, with a batch-wise distribution-aware calibration factor.
However, as widely reported in the prompt-based learning literature, these methods often suffer from performance degradation when applied to domains for which the prompts were not trained~\cite{Yu2023, Wang2022a, Zhou2022a}.
Unlike existing approaches, our method dynamically balances the use of prefixes and adapters to leverage the strengths of both. 
\section{Preliminaries}
\subsection{Continual Learning Protocol for VLMs}
Following~\cite{Tang2024, Yu2024, Zheng2023},
we consider a collection of \( T \) tasks, denoted by \(\{\mathcal{T}^t\}_{t=1}^{T}\). Each task \(\mathcal{T}^t\) consists of a dataset \( D_t \) and a set of class names \( C_t \). The class names in \( C_t \) are paired with manually crafted prompts and fed into the text encoder.
The dataset \( D_t \), consisting of \( N_t \) samples, is represented as \( D_t = \{(I_{tn}, y_{tn})\}_{n=1}^{N_t} \), where \( I_{tn} \) is the input image and \( y_{tn} \) is its corresponding one-hot label from \( C_t \).
In CL, models incrementally learn by sequentially processing each task \(\mathcal{T}^t = \{D_t, C_t\}\). 
The primary objective of CL in VLMs is to maintain strong performance on both learned and unseen tasks, with the latter evaluated for zero-shot capabilities after each task to measure a new type of forgetting that arises in the context of CL for VLMs, known as forward forgetting.
We consider two relevant benchmarks: Multi-domain Task Incremental Learning (MTIL)~\cite{Zheng2023} and Open-Domain Continual Learning (ODCL-CIL)~\cite{Li2024}, where both domain and class distributions change across tasks. In MTIL, the model has access to the task ID during evaluation, whereas ODCL-CIL requires classification over all learned classes without task ID information.

\subsection{Prompt-Based Continual Learning}
Prompt-based methods are widely used in CL to adapt transformer-based pretrained models that process data samples in token units~\cite{Wang2022c, Wang2022b, Smith2023, Wang2022a, Jung2023}. 
These methods adjust the model by introducing prompt tokens of length \(L\), consisting of learnable vectors of dimension \(d\), while keeping the model's backbone frozen.
For each task \(t\), a unique set of vectors \(P_t \in \mathbb{R}^{L \times d}\) is learned, constructing a pool \(\{P_1, P_2, \ldots, P_T\}\) that spans \(T\) tasks.
At inference time, the prompt \( P_t \) is selected from the pool and attached to the pretrained model to restore task-specific knowledge~\cite{Tang2024}, allowing input tokens \( X \in \mathbb{R}^{m \times d} \) to absorb task-specific information through the attention mechanism by modifying the inputs of multi-head self-attention layers~\cite{Xing2022, Wang2022b}.
Among prompt-based methods, prefix-tuning~\cite{Li2021Prefix} learns a pair of task-specific prompts \( P_K, P_V \in \mathbb{R}^{L \times d} \) instead of a single \( P \), and separately projects them to the key and value in the attention mechanism.

Specifically, for the \(i\)-th head, input tokens and prefix are first transformed into query \(Q_i\), key \(K_i\), and value \(V_i\) by corresponding projection matrices \(W_i^Q\), \(W_i^K\), \(W_i^V\) and their respective biases. With the input token \(X\), this can be formulated as follows: 
\begin{equation}
\begin{aligned}
Q_i &= X W_i^Q + \mathbf{1}_n\, b_i^Q \in \mathbb{R}^{n \times \frac{d}{h}}, \\
K_i &= \begin{bmatrix} 
K_{X_i} \\ 
K_{P_i} 
\end{bmatrix} = 
\begin{bmatrix} 
X W_i^K + \mathbf{1}_n\, b_i^K \\ 
P_K W_i^K + \mathbf{1}_L\, b_i^K
\end{bmatrix} \in \mathbb{R}^{\left(n + L\right) \times \frac{d}{h}}, \\
V_i &= \begin{bmatrix} 
V_{X_i} \\ 
V_{P_i} 
\end{bmatrix} = 
\begin{bmatrix} 
X W_i^V + \mathbf{1}_n\, b_i^V \\ 
P_V W_i^V + \mathbf{1}_L\, b_i^V
\end{bmatrix} \in \mathbb{R}^{\left(n + L\right) \times \frac{d}{h}}.
\end{aligned}
\end{equation}
The self-attention computes the attention scores \(S_i\) using the inner product of query and key \(Q_i(K_i)^\top\):
\begin{equation}
\begin{aligned}
S_i 
&= \left[ S_{XX}^{(i)},\, S_{XP}^{(i)} \right] 
= \left[ Q_i \,(K_{X_i})^\top,\; Q_i \,(K_{P_i})^\top \right].
\end{aligned}
\label{eq:attention_scores}
\end{equation}
Here, \( S_{XX}^{(i)} \) represents the attention scores among the input tokens, while \( S_{XP}^{(i)} \) represents the attention scores between the input tokens and the prefixes. Next, these scores are normalized to attention weights using a Softmax function with temperature $\tau$, which are then used to weight the corresponding value matrix \(V_i\):
\begin{equation}
h_i = \text{softmax}\left(\frac{S_i}{\sqrt{d/h}}\right)V_i \in \mathbb{R}^{m \times \frac{d}{h}}.
\label{eq:softmax_normalization}
\end{equation}
Finally, the outputs from all heads are concatenated and projected through the projection matrix \(W_o\). 
\begin{figure*}[t!]
\begin{center}
\resizebox{1.0\linewidth}{!}{
\includegraphics[width=\textwidth]{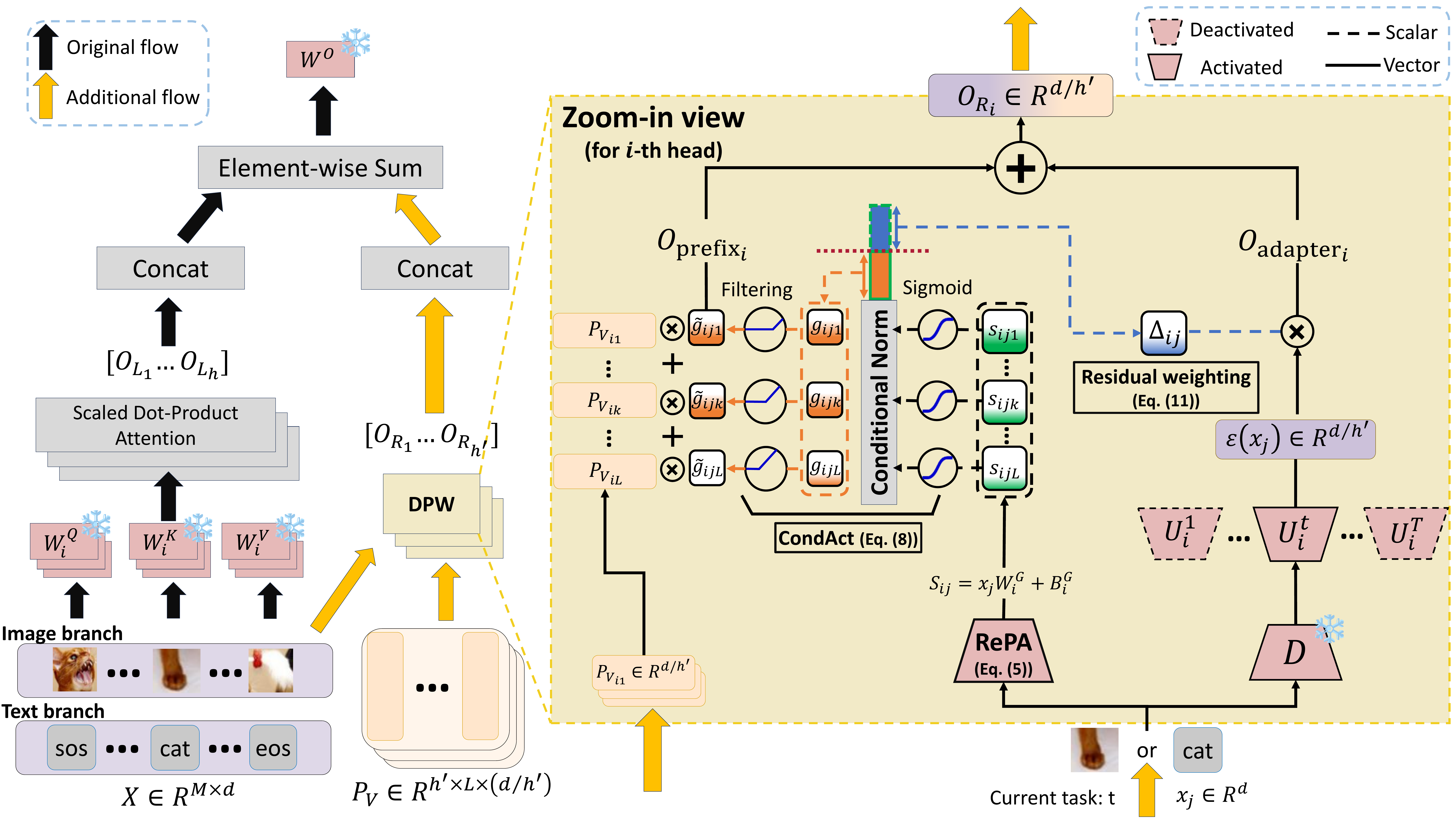}
}
\vspace{-15.0pt}
\caption{Overall framework of the proposed method. The right side of the figure shows how each input token assigns weights to both the prefixes and the adapter. Specifically, we compute prefix scores for each input token using a RePA module, and then convert those scores into weights with the CondAct module. We also set an upper limit on the total weight that can be assigned to the prefixes. Any weight beyond this limit is applied to the adapter’s output. This design enables token-level weighting for the adapter without the need for a router.}
\vspace{-20.0pt}
\label{fig:fremework}
\end{center}
\end{figure*}

\section{Methods}
The main goals of our method are two-fold.
First, we aim to ensure that each input token receives a proper amount of information from the prefixes and adapters throughout CL tasks. 
Second, we aim to establish an effective synergy between the prefixes and adapters.
Our method addresses both motivations within a unified framework. 
Specifically, we first introduce a gating module that assigns refined weights to the prefixes. Then, guided by these weights, tokens that require additional adaptation receive information from the adapter. Fig.~\ref{fig:fremework} provides an overview of the proposed framework.

\subsection{Reparameterized Prefix Attention}
\begin{figure}[t!]
\begin{center}
    \includegraphics[width=0.73\columnwidth]{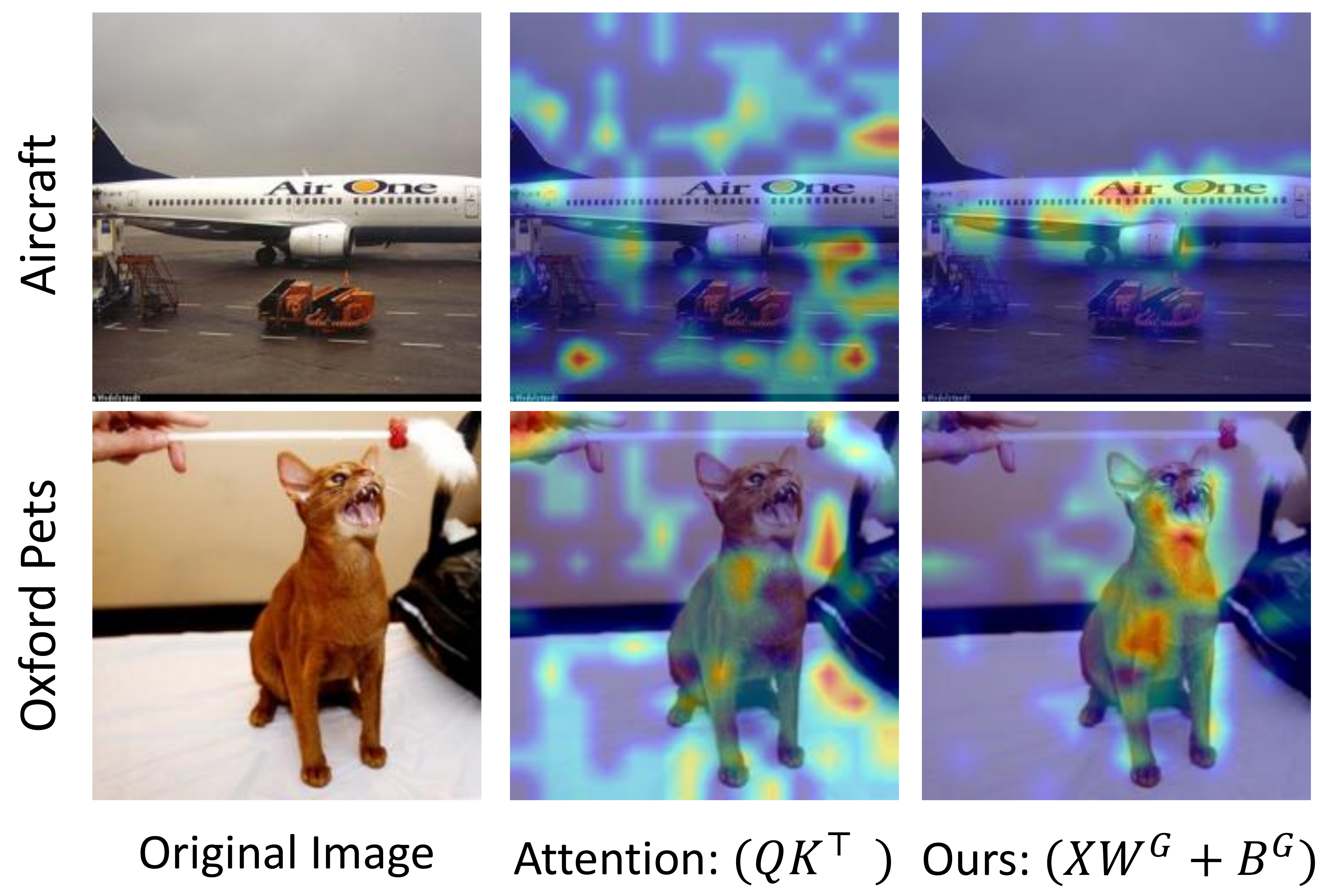}
    \vspace{-10pt}
    \caption{Comparison of prefix score maps between attention and our method on the Aircraft and Oxford Pets images. Our approach produces more task-relevant scores.}
    \label{fig:attn_score}
\vspace{-27pt}
\end{center}
\end{figure}

In conventional prefix-tuning for CL, the learnable prefix key $P_K$ is appended for each task and projected onto the pretrained key distribution through a key projection matrix.
The $P_K$ is then optimized to align with the pretrained query distribution of the corresponding task, determining the attention scores that weight the prefix value $P_V$.
However, because the pretrained model typically lacks sufficient task-specific knowledge for new tasks~\cite{Zheng2023}, these attention scores often fail to accurately capture the task relevance of input tokens, as illustrated in Fig.~\ref{fig:attn_score}.
Consequently, the model fails to effectively regulate the amount of information injected per token, which has been shown to impair downstream task performance~\cite{Fu2022, Zhou2024}.

A straightforward way to address this issue is to learn a separate query-key projection matrix for each task, dedicated exclusively to computing the attention scores of the prefix.
Although this approach improves performance in practice, the large number of additional parameters, together with the limited data in downstream tasks, often leads to suboptimal results.
Instead, we propose a reparameterized prefix attention \textbf{(RePA)}, which simplifies the query-key dot product process into a single affine transformation to compute the prefix scores.
Specifically, for the \(i\)-th attention head, the original attention module computes the prefix score through \(S_{XP}^{(i)} = Q_i (K_{P_i})^\top = (XW_i^Q + \mathbf{1}_m b_Q^\top)(P_K W_i^K + \mathbf{1}_L b_K^\top)^\top\). 
This expression can be rewritten by separating the terms that depend on the input token matrix $X$ from those that remain constant, yielding the following equation:
\begin{equation}    
\begin{split}
S_{XP}^{(i)} &= X \Bigl[\, W_i^Q (P_K W_i^K)^\top + W_i^Q b_K\, \mathbf{1}_L^\top \Bigr] \\
&\quad + \Bigl[\, \mathbf{1}_m\,b_Q^\top (P_K W_i^K)^\top + \left(b_Q^\top b_K\right) \mathbf{1}_{m \times L} \Bigr].
\end{split}
\end{equation}
We can group the terms into two matrices, arriving at RePA:
\begin{equation}
S_{XP}^{(i)} = X\, W_i^G + B_i^G.
\label{eq:affine}
\end{equation}
Note that training \(W_i^G\) and \(B_i^G\) effectively unifies the original attention projection parameters and the learnable prefix key \(P_K\) into a single reparameterized form~\cite{oymak2023}:
\begin{align}
W_i^G 
&\approx
W_i^Q\,(P_K W_i^K)^\top + W_i^Q\,b_K\,\mathbf{1}_L^\top, \\
B_i^G
&\approx
\mathbf{1}_m\,b_Q^\top\,(P_K W_i^K)^\top + (b_Q^\top b_K)\,\mathbf{1}_{m\times L}.
\label{eq:affine_components}
\end{align}

By training these composite parameters, $W_i^G$ and $B_i^G$, we can better capture task relevance by learning task-specific projections applied exclusively to the prefixes, while preserving pretrained knowledge by leaving the original projection matrices unchanged. 
Consequently, as illustrated in Fig.~\ref{fig:attn_score}, our RePA assigns higher attention scores to tokens that are directly relevant to each task, thereby facilitating more effective adaptation to downstream tasks~\cite{Fu2022, Zhou2024}.

\subsection{Conditional Activation}
Applying the above method, we obtain an enhanced score matrix \(S_{XP}^{(i)}\). 
However, prior work~\cite{Tang2024} has demonstrated that concatenating \(S_{XP}^{(i)}\) with the original attention score \(S_{XX}^{(i)}\) before applying the softmax function disrupts the pretrained knowledge encoded in \(S_{XX}^{(i)}\), leading to significant degradation in zero-shot performance.
Alternatively, DIKI~\cite{Tang2024} proposes applying a separate softmax to \(S_{XP}^{(i)}\).
While this approach avoids interference with the pretrained scores, it imposes a fixed total weight across all input tokens, hindering dynamic adjustment of the additional information for each token, thereby limiting token-wise modulation.

To overcome these limitations, we introduce Conditional Activation (\textbf{CondAct}), a novel alternative to the conventional softmax activation.
CondAct processes the score matrix $S_{XP}$ through two conditional operations, conditional normalization and conditional filtering, applied after a sigmoid function $\sigma(\cdot)$.
Let \(s_{ijk}\) denote the score between the \(j\)-th input token and the \(k\)-th prefix, corresponding to the \((j,k)\)-th element of the score matrix \(S_{XP}^{(i)}\). 
The corresponding prefix weight \(\widetilde{g}_{ijk}\) is computed via the following procedure:
\begin{subequations}\label{eq:condact}

\noindent\text{(Step 1: Conditional Norm)}
\begin{align}
g_{ijk} &=
\begin{cases}
\displaystyle \frac{\sigma(s_{ijk})}{\sum_{k=1}^{L}\sigma(s_{ijk})}, 
   & \text{if } \sum_{k=1}^{L}\sigma(s_{ijk}) \ge 1, \\[12pt]
\sigma(s_{ijk}), 
   & \text{otherwise}.
\end{cases}
\label{eq:condact:condnorm}
\end{align}

\noindent\text{(Step 2: Conditional Filtering)}
\begin{align}
\widetilde{g}_{ijk}
&= g_{ijk}\,\cdot\,\mathbb{I}\bigl(g_{ijk} \ge \mathrm{cutoff}\bigr).
\label{eq:condact:filtering}
\end{align}
\end{subequations}
Here, $\mathbb{I}(\cdot)$ is an indicator function that returns 1 if the condition is met and 0 otherwise.

The conditional normalization step in Eq.~\eqref{eq:condact:condnorm} enables the total sum of prefix weights, $\sum_{k=1}^{L} g_{ijk}$, to flexibly range up to one---allowing the model to dynamically adjust the influence of each input token.
This upper bound prevents task-specific information from becoming overly dominant, a common issue when using a sigmoid function alone~\cite{Ramapuram2025}.
Consequently, CondAct mitigates forward forgetting in VLMs and enhances zero-shot generalization in CL tasks.

Next, the conditional filtering step in Eq.~\eqref{eq:condact:filtering} removes prefixes that are less relevant to the current task.
Empirically, we observe that the attention scores between the [CLS] token and each prefix follow a Gaussian distribution. 
Leveraging this property, we determine the cutoff value dynamically for each sample.
Specifically, for each prefix, we compute the mean and variance of its attention scores with the [CLS] token across the training data of task $t$ and model its distribution as \( \mathcal{N}(\mu_t, \sigma^2_t) \).
Given a test sample, we evaluate the log probability density of the observed attention score $s_{i,\text{cls},k}$ under this distribution and transform it with a sigmoid function to obtain a likelihood score, as in~\cite{Tang2024}.
The cutoff for the \(k\)-th prefix is defined as one minus this score: 
\begin{equation}
\text{cutoff}_{ijk} = 1 - \sigma\Bigl(\log \varphi(s_{i,\text{cls},k}; \mu_t, \sigma^2_t)\Bigr).
\label{eq:cutoff_nontxt}
\end{equation}
If the likelihood score is sufficiently high, we set the cutoff to zero so that the corresponding prefix fully contributes to the output.
Since each prefix exhibits distinct activation patterns across input tokens, this prefix-wise adaptive cutoff enables token-wise dynamic filtering conditioned on task-specific likelihoods, allowing the model to selectively suppress irrelevant prefixes. 
For the \(i\)-th head, the additional information derived from the prefixes is computed as follows:
\begin{equation}
O_{\text{prefix}_i} = \widetilde{G}_i \, P_{V_{i}} \quad \text{with} \quad \widetilde{G}_i = \left( \widetilde{g}_{ijk} \right)_{j,k} \in \mathbb{R}^{m \times L}.
\label{eq:prompt_output}
\end{equation}

\subsection{Residual Weighting Mechanism}

\begin{table*}[t!]
\centering
\begin{minipage}[t]{0.535\textwidth}
\centering
\resizebox{1.0\linewidth}{!}{%
    \begin{tabular}{lcc|ccc| c}
      \toprule
      \textbf{Method} & \textbf{Extra data} & \textbf{Params.} & \textbf{Trans.} & \textbf{Avg.} & \textbf{Last} & \textbf{Mean} \\
      \midrule
      Zero-shot           & - & – & 69.4 & 65.3 & 65.3 & 66.7 \\
      \midrule
      ZSCL~\cite{Zheng2023}           & $\checkmark$ & 149.6 M & 68.1 & 75.4 & 83.6 & 75.7 \\
      DIKI~\cite{Tang2024}            & $\times$     & 1.8 M & 68.7 & 76.3 & 85.1 & 76.7 \\
      MoE-Adapter~\cite{Yu2024}       & $\times$     & 59.6 M & 68.9 & 76.7 & 85.0 & 76.9 \\
      GIFT~\cite{GIFT}                & $\checkmark$ & 149.6 M & 69.3 & 77.3 & 86.0 & 77.5 \\
      \midrule
      \rowcolor{gray!20}
      Ours\textdagger     & $\times$     & 4.6 M & \underline{70.0} & \underline{78.6} & \underline{87.6} & \underline{78.7} \\
      \rowcolor{gray!20}
      Ours                & $\times$     & 30.8 M & \textbf{70.4} & \textbf{79.3} & \textbf{88.3} & \textbf{79.3} \\
      \bottomrule
    \end{tabular}%
  }\vspace{-7.0pt}
\captionof{table}{Comparison of various SOTA methods on MTIL Order I benchmark in terms of ``Transfer'', ``Average'', and ``Last'' scores (\%). Best and second-best results are highlighted in \textbf{bold} and \underline{underline}, respectively.}
\label{tab:mtli-order1}
\end{minipage}
\hfill
\begin{minipage}[t]{0.43\textwidth}
\centering
\resizebox{1.0\linewidth}{!}{%
    \begin{tabular}{l | c c c| c}
    \toprule
    \textbf{Method} & \textbf{Trans.} & \textbf{Avg.} & \textbf{Last} & \textbf{Mean} \\
    \midrule
    ZSCL~\cite{Zheng2023}             & 68.0 & 71.8 & 77.6 & 72.5 \\
    MoE-Adapter~\cite{Yu2024}         & 69.1 & 66.2 & 66.9 & 67.4 \\
    CoLeCLIP~\cite{Li2024}            & 68.8 & 73.7 & 79.7 & 74.1 \\
    DPeCLIP~\cite{Lu2024}             & 69.1 & 76.1 & 84.6 & 76.6 \\
    \midrule
    \rowcolor{gray!20}
     Ours\textdagger  & \underline{70.0} & \underline{77.9} & \underline{85.8} & \underline{77.9} \\
     \rowcolor{gray!20}
     Ours & \textbf{70.4} & \textbf{78.6} & \textbf{86.6} & \textbf{78.5} \\
    \bottomrule
  \end{tabular}
  }
\vspace{-7.0pt}
\captionof{table}{Comparison of various SOTA methods on ODCL-CIL benchmark in terms of ``Transfer,'' ``Average,'' and ``Last'' scores (\%).}
\label{tab:mcil-order1}
\end{minipage}

\vspace{-16.0pt}
\end{table*}

Our proposed CondAct dynamically adjusts the total weight of the prefixes for each token, constraining it within an upper limit of one to preserve zero-shot performance.
However, strictly enforcing this limit may restrict further improvements, as tokens with higher task relevance often require stronger adaptation.
To address this, we introduce the Residual Weighting Mechanism \textbf{(RWM)}, which selectively applies the adapter output to tokens that demand additional adaptation.
For parameter efficiency, we employ LoRA as the adapter~\cite{Yu2024} and follow the structure proposed in~\cite{Tian2024HydraLoRA}, where a shared down-projection matrix $D$ is used across tasks and a task-specific up-projection $U_t$ is assigned to each task $t$.
To further adapt this design for CL, we freeze the shared projection, initializing it with the top-$k$ left singular vectors of the value projection matrix $W_i^V$ as~\cite{Meng2024PiSSA, Wang2024MiLoRA}.
This constraint ensures that the adapter is fine-tuned within the row space of $D$, effectively leveraging pretrained knowledge to complement the learned prefixes~\cite{Liang2024}.
The adapter's output, denoted as \(\mathcal{E}^t_i(X)\), is then weighted using RWM:
\begin{equation}
\label{eq:adapter_delta}
\begin{array}{c}
O_{\mathrm{adapter}_i} = \Delta_i \odot \mathcal{E}^t_i(X), \\[8pt]
\Delta_i = \mathrm{concat}\bigl[\Delta_{ij}\bigr]_{j=1}^m, \\[6pt]
\Delta_{ij} = \max\!\Bigl(0,\,\sum_{k=1}^L \sigma(s_{ijk}) - 1\Bigr),
\end{array}
\end{equation}
where \(\odot\) denotes element-wise multiplication. 

If the sum of prefix weights, $\sum_{k=1}^{L}\sigma(s_{ijk})$, does not exceed one, the corresponding element of $\Delta_i$ becomes zero, meaning that the adapter does not contribute to that token.
In this way, RWM leverages the residual prefix weights as adaptive scaling factors, enabling each token to receive additional information proportional to its remaining adaptation demand.
Through this combination of prefixes and adapters, we achieve a balance between the aggressive modification introduced by the prefix and the conservative update performed by the adapter.
For each head \(i\), the output of the proposed DPW module, denoted as $O_{R_i}$ is computed as follows:
\begin{equation}
O_{R_i} = O_{\text{prefix}_i} + O_{\text{adapter}_i}.
\label{final_formula}
\end{equation} 
\section{Experiments}

\begin{table}[t!]
\centering
\resizebox{0.76\columnwidth}{!}{%
    \begin{tabular}{ccc|ccc}
      \toprule
      RePA & CondAct & RWM & Trans. & Avg. & Last \\
      \midrule
      - & - & - & 68.1 & 76.6 & 85.9 \\
      \midrule
      \checkmark & - & - & 68.0 & 76.8 & 86.4  \\
      - & \checkmark & - & 69.5 & 77.8 & 86.8  \\
      \checkmark & \checkmark & - & 69.9 & 78.9 & 87.9\\
      \rowcolor{gray!20}
      \checkmark & \checkmark & \checkmark & \textbf{70.4} & \textbf{79.3} & \textbf{88.3} \\
      \bottomrule
    \end{tabular}%
  }
\vspace{-6.0Pt}
\caption{Ablation study on the MTIL benchmark evaluating each module's contribution.}
\label{tab:ablation}
\vspace{-16.0pt}
\end{table}

\subsection{Experimental Setting}
\textbf{Benchmarks.} We evaluate our method in two domain-class incremental settings: Multi-domain Task Incremental Learning (MTIL)~\cite{Zheng2023} and Open-Domain Continual Learning (ODCL-CIL)~\cite{Li2024}. MTIL uses task IDs for task-specific classification, while ODCL-CIL classifies across all seen classes without task IDs. Both benchmarks include 11 datasets from various domains, covering 1201 classes.

\vspace{2pt}\noindent\textbf{Metrics.} 
We evaluate our method on both the MTIL and ODCL-CIL benchmarks using the metrics proposed in~\cite{Zheng2023, Li2024}: \textit{Transfer}, \textit{Avg.}, and \textit{Last}.
The \textit{Transfer} score measures a model's ability to generalize to unseen data through zero-shot evaluation, and it is used to quantify forward forgetting in the context of CL for VLMs~\cite{Zheng2023}.
Because task IDs are used when evaluating zero-shot capabilities in both benchmarks, the \textit{Transfer} score is identical across MTIL and ODCL-CIL.
The \textit{Last} score represents the average performance across all tasks at the end of CL, whereas the \textit{Avg.} score measures the average accuracy across all datasets and training stages.

\vspace{2pt}\noindent\textbf{Compared Methods.} 
We compare our method against various state-of-the-art (SOTA) approaches, including prompt-based, adapter-based, and full fine-tuning methods.
For prompt-based learning, we compare CoLeCLIP~\cite{Li2024}, DIKI~\cite{Tang2024} and DPeCLIP~\cite{Lu2024}.
For adapter-based learning, we compare MoE-Adapters~\cite{Yu2024}. For full fine-tuning, we compare ZSCL~\cite{Zheng2023} and GIFT~\cite{GIFT}.
In addition to our default model (denoted Ours), we also introduce a parameter-efficient variant, Ours\textdagger, which reduces the number of trainable parameters.
Ours\textdagger computes the prefix score for each head using only its corresponding sub-dimension (i.e., $d/h^{\prime}$) instead of the full embedding, reducing the number of parameters by a factor of $h^{\prime}$. We further apply LoRA with a lower rank for additional efficiency.
All experiments follow the baseline setup~\cite{Tang2024}, with details provided in the Appendix.

\begin{figure*}[t!]
\centering
\begin{minipage}[t]{0.465\textwidth}
\centering
\includegraphics[width=\linewidth]{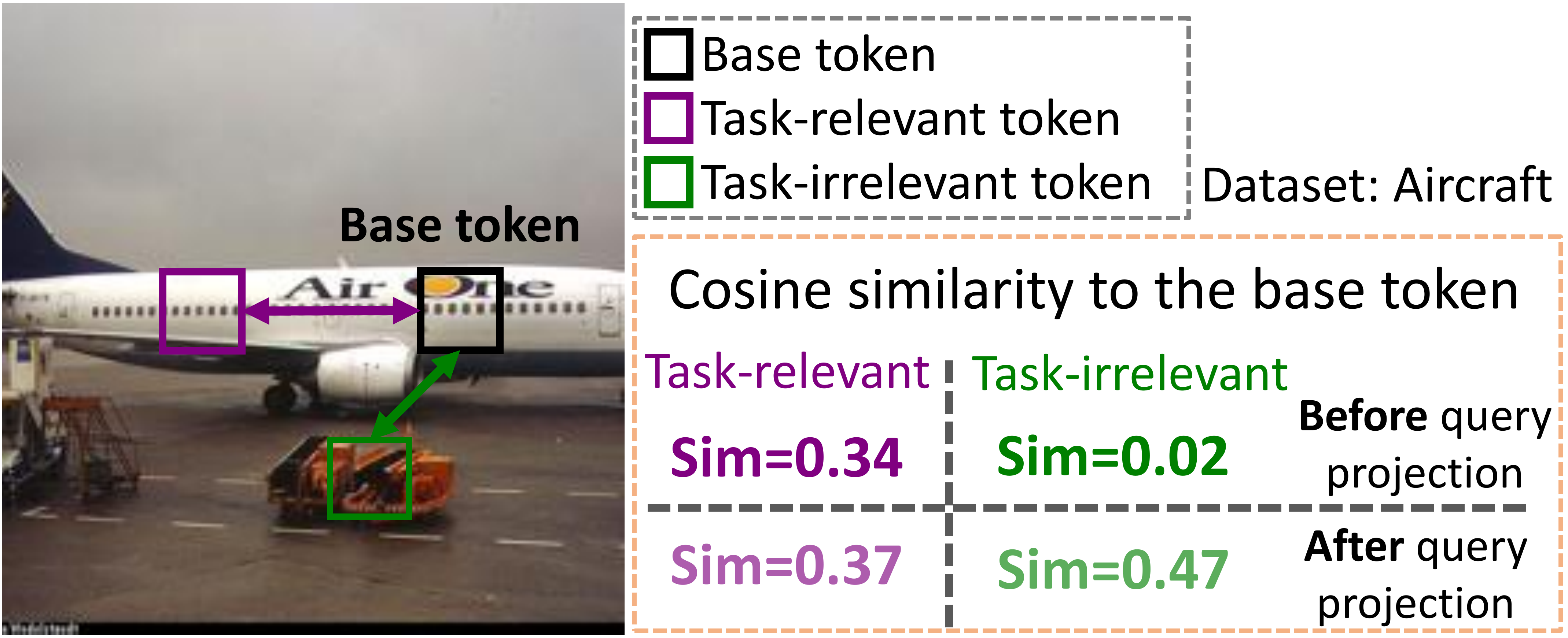}
\vspace{-20.0pt}
\captionof{figure}{Cosine similarities comparison between the 
      \textbf{\textcolor{black}{base token}} and both the 
      \textbf{\textcolor[HTML]{800080}{task-relevant}} and 
      \textbf{\textcolor[HTML]{008000}{task-irrelevant}} tokens, 
      shown before and after query projection.}
\label{fig:patch_sim}
\end{minipage}
\hfill
\begin{minipage}[t]{0.465\textwidth}
\centering
\includegraphics[width=\linewidth]{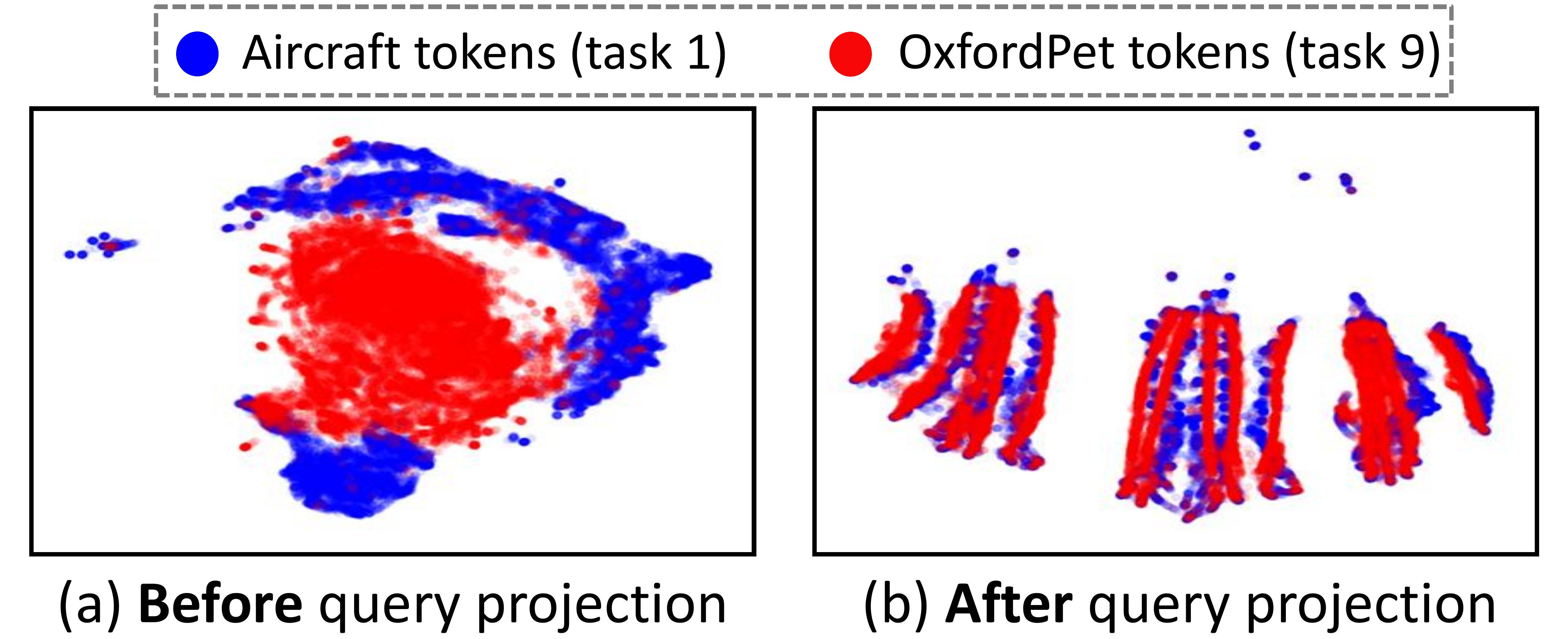}
\vspace{-20.0pt}
\captionof{figure}{UMAP visualization of token embeddings from the \textbf{\textcolor[HTML]{0000FF}{Aircraft}} and \textbf{\textcolor[HTML]{F70707}{OxfordPet}} datasets, illustrating their distributions before and after applying query projection.}
\label{fig:umap_sim}
\end{minipage}
\vspace{-10.0pt}

\end{figure*}

\begin{table*}[t!]
\centering
\begin{minipage}[t]{0.34\textwidth}
\centering
\resizebox{\linewidth}{!}{
  \begin{tabular}{c|c|ccc}
      \toprule
      Method & Params. & Trans. & Avg. & Last \\
      \midrule
      $QK^\top$ (freeze $W^Q$, $W^K$) & 2.7 M & 68.7 & 76.4 & 85.2 \\
      $QK^\top$ (train $W^Q$, $W^K$) & 228.0 M & 68.9 & 76.8 & 85.9 \\
      $QW_G + B_G$ & 30.8 M & 69.7 & 78.6 & 87.6 \\
      \midrule
      \rowcolor{gray!20}
      $XW_G + B_G$ (Ours) & 30.8 M & \textbf{70.4} & \textbf{79.3} & \textbf{88.3} \\
      \bottomrule
    \end{tabular}%
}
\vspace{-7.0pt}
\captionof{table}{Comparison between RePA and the traditional attention mechanism.} 
\label{tab:RePA_analysis}
\end{minipage}
\hfill
\begin{minipage}[t]{0.33\textwidth}
\centering
\resizebox{\linewidth}{!}{
  \begin{tabular}{ccc|ccc}
      \toprule
      Sigmoid & CondNorm & Filtering & Trans. & Avg. & Last \\
      \midrule
      - & - & - & 68.0 & 76.8 & 86.4 \\
      \midrule
      \checkmark & - & - & 68.6 & 78.3 & 88.2 \\
      \checkmark & \checkmark & - & 69.9 & 79.0 & 88.2 \\
      \rowcolor{gray!20}
      \checkmark & \checkmark & \checkmark & \textbf{70.4} & \textbf{79.3} & \textbf{88.3} \\
      \bottomrule
    \end{tabular}%
}
\vspace{-7.0pt}
\captionof{table}{Ablation study evaluating the impact of different components within the CondAct.}
\label{tab:CondAct_analysis}
\end{minipage}
\hfill
\begin{minipage}[t]{0.295\textwidth}
\centering
\resizebox{\linewidth}{!}{
  \begin{tabular}{c|c|ccc}
    \toprule
    Method & Params. & Trans. & Avg. & Last \\
    \midrule
    Baseline & 19.8 M & 69.9 & 78.9 & 87.9\\
    Prefix length $\times2$ & 39.6 M & 69.5 & 78.4 & 87.6 \\
    \midrule
    Learnable router & 32.7 M & 69.9 & 79.0 & 88.1 \\
    \rowcolor{gray!20}
    RWM (Ours) & 30.8 M & \textbf{70.4} & \textbf{79.3} & \textbf{88.3} \\
    \bottomrule
    \end{tabular}%
}
\vspace{-7.0pt}
\captionof{table}{Comparison between the RWM and traditional routing mechanism.}
\label{tab:RWM_analysis}
\end{minipage}
\vspace{-15.0pt}

\end{table*}

\subsection{Main Results.}

\textbf{Performance on MTIL.} 
Tab~\ref{tab:mtli-order1} presents a comparison between the proposed method and several SOTA approaches on the MTIL benchmark.
For the sequence of training tasks, we follow the predefined Order I configuration described in~\cite{Zheng2023}, where tasks are arranged alphabetically.
Detailed results, including Order II, are provided in the Appendix.
Overall, the results show that both Ours and Ours\textdagger~ outperform competing methods across all evaluation metrics. 
In particular, they surpass the second-best approach, GIFT~\cite{GIFT}, which performs full fine-tuning of the entire CLIP model using additional data generated by an auxiliary diffusion model.
Compared to PEFT methods, such as MoE-Adapter~\cite{Yu2024} and DIKI~\cite{Tang2024}, our approach consistently achieves superior performance while maintaining a similarly low number of trainable parameters.
The strong performance on both \textit{Transfer} and \textit{Last} scores indicates that our method not only adapts effectively to new tasks but also preserves previously learned knowledge, underscoring its suitability for CL.

\vspace{2pt}\noindent\textbf{Performance on ODCL-CIL.} 
We evaluate our method on class-incremental setting using the ODCL-CIL benchmark, as presented in Tab.~\ref{tab:mcil-order1}. 
Compared with the MTIL benchmark, the ODCL-CIL scenario is more challenging because task IDs are unavailable during inference.
Following the baseline method~\cite{Tang2024}, we adopt its task-identification strategy, which estimates the likelihood of the current image feature under each task distribution and selects the parameters associated with the highest-likelihood task.
To avoid potential advantages from batch voting~\cite{batchvoting2024}, we further evaluate our method using both the default batch size (e.g., 256) and a batch size of one, reporting the lower score.
Across all evaluation metrics, our method consistently achieves SOTA performance in this class-incremental setting.

\subsection{Analysis}

\noindent\textbf{Ablation Study.} 
Our framework consists of three components: \textbf{RePA} to compute prefix scores (Eq.~\eqref{eq:affine}), \textbf{CondAct} to transform the prefix scores into weights (Eq.~\eqref{eq:condact}), and \textbf{RWM} to incorporate the adapter (Eq.~\eqref{eq:adapter_delta}). 
We evaluate each module on the MTIL benchmark in Tab.~\ref{tab:ablation}.
First, by replacing the traditional query-key dot product with the RePA, we obtain a refined prefix score matrix \(S_{XP}\), leading to an improvement in the \textit{Last} score. 
However, this component alone cannot overcome the inherent limitation of softmax, which forces scores to sum to one and thus restricts the optimal assignment of weight per token.
As shown in the third row, replacing softmax with our CondAct improves overall performance by dynamically controlling the amount of information added to each token, and combining it with the refined scores from RePA yields additional gains by better capturing task-specific characteristics in the scoring process.
This token-wise modulation not only improves performance on the trained tasks, but also facilitates effective knowledge transfer across tasks during CL and mitigates forward forgetting, as evidenced by the improved \textit{Transfer} scores.
Moreover, by leveraging the weights assigned to each prefix, our RWM enables adapters to add information only to tokens that are not sufficiently adapted by the prefixes, thereby effectively complementing the prefixes and leading to notable improvements in both the \textit{Last} and \textit{Transfer} scores.

\vspace{2pt}\noindent\textbf{Limitations of Pretrained Attention Projections in CL.} 
To further investigate the limitations of existing attention mechanisms in prefix-tuning for CL, we conduct a comparative analysis of token embeddings before and after the query projection. 
As shown in Fig.~\ref{fig:patch_sim}, we extract tokens from a single image and label those from object regions (e.g., the body of an airplane) as task-relevant and those from background regions as task-irrelevant, then compute cosine similarities (i) among task-relevant tokens (purple arrow) and (ii) between task-relevant and task-irrelevant tokens (green arrow).
Specifically, we find that before the query projection, the similarity between task-relevant and task-irrelevant tokens is merely 0.02.
After the projection, however, this similarity increases substantially to 0.47, exceeding even the similarity among task-relevant tokens.
This indicates that the query projection disrupts the model's ability to maintain distinctions between task-relevant and task-irrelevant features. 
Consequently, as shown in Fig.~\ref{fig:attn_score}, prefix struggles to route attention to the appropriate tokens.
Notably, this effect is not limited to individual samples; it also appears at the task level in our CL setting.
Fig.~\ref{fig:umap_sim} illustrates UMAP visualizations of tokens from multiple samples across two tasks. 
Before the projection, tokens from different tasks are clearly separated. 
However, after projection, they collapse into overlapping regions, revealing diminished task separability.
Such reduced separability increases task interference and weakens effective knowledge transfer across CL tasks~\cite{Hiratani2024}.
Overall, these findings highlight inherent limitations in approaches that rely on pretrained attention projections~\cite{Tang2024, Lu2024, Wang2022b, Wang2022c}.

\vspace{2pt}\noindent\textbf{Analysis of RePA.} 
To evaluate whether RePA can effectively replace the conventional attention mechanism, we conduct experiments summarized in Tab.~\ref{tab:RePA_analysis}.
As shown in the second row, re-training the attention projection matrices leads to improved CL performance.
However, this approach substantially increases the number of learnable parameters to 228.0M, while the performance gain remains significantly lower than that achieved by RePA.
In the third row, we preserve the original query projection process of the attention mechanism and apply RePA on top of it. 
Similar to the previous case, the performance remains inferior to that obtained when RePA is directly applied to the input tokens.
These results highlight the intrinsic limitations of the attention mechanism and demonstrate that RePA serves as an effective and efficient alternative.

\begin{figure}[t!]
\vspace{-5pt}
\begin{center}
    \includegraphics[width=0.88\columnwidth]{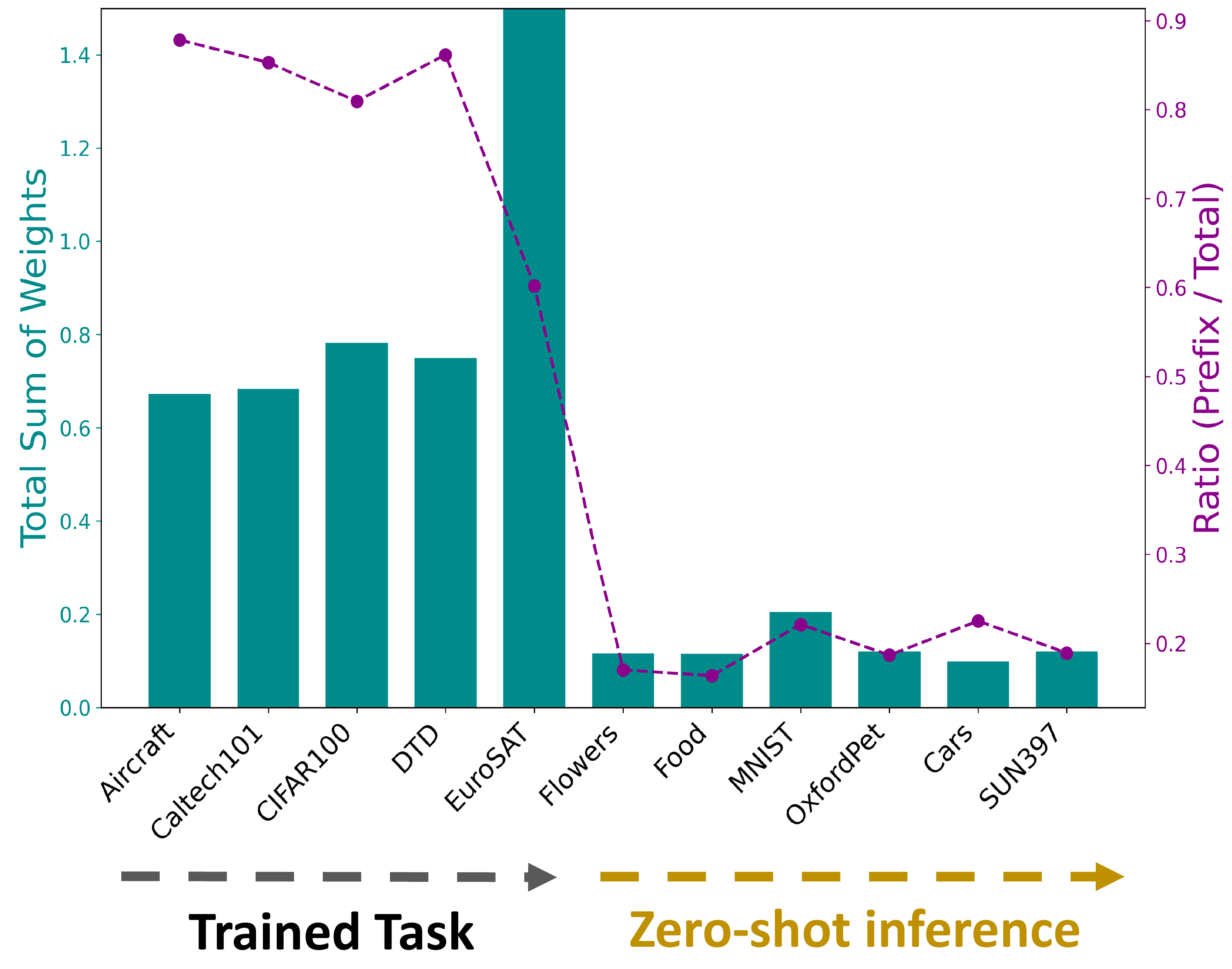}
    \vspace{-10pt}
    \caption{Weight distribution after training on five tasks. \textbf{\textcolor[HTML]{008B8B}{Green bars}} show the sum of prefixes and adapters weights, while \textbf{\textcolor[HTML]{8B008B}{purple lines}} indicate the ratio of prefix weights to the total.}
    \label{fig:weights_analysis}
\vspace{-27.5pt}
\end{center}
\end{figure}

\vspace{2pt}\noindent\textbf{Analysis of CondAct.} 
To assess the contribution of each component in the proposed CondAct mechanism, we conduct an ablation study summarized in Tab.~\ref{tab:CondAct_analysis}.
Consistent with recent findings~\cite{Le2024}, which report improved performance when combining softmax and sigmoid activations in prefix tuning, we observe that incorporating a sigmoid activation into RePA also yields notable gains.
However, the sigmoid function tends to produce overly large total weights for input tokens~\cite{Ramapuram2025}, potentially distorting their pretrained representations and degrading \textit{Transfer} performance.
Our conditional normalization module~\eqref{eq:condact:condnorm} mitigates this issue by maintaining the total prefix information within a stable range, leading to a substantial improvement in \textit{Transfer} accuracy.
Furthermore, the conditional filtering mechanism~\eqref{eq:condact:filtering}, which leverages the distribution of prefix scores to assign prefix-wise cutoffs, enables dynamic and token-wise selection of informative prefixes.
This adaptive filtering further enhances the \textit{Transfer} score by refining how prefix information is selectively utilized across tokens.

\vspace{2pt}\noindent\textbf{Analysis of RWM.}
To assess the effectiveness of our RWM, we conduct experiments summarized in Tab.~\ref{tab:RWM_analysis}.
We first observe that simply increasing the number of learnable parameters does not yield better performance. 
As shown in the second block, doubling the prefix length degrades performance, highlighting the limitation of performance gains achievable through prefix alone. 
The third block introduces an adapter with a learnable router, but the resulting improvements remain marginal, suggesting that the adapter is ineffective without an appropriate weighting mechanism.
In contrast, our proposed RWM module enables more effective weighting, leading to consistent improvements in both the \textit{Transfer} and \textit{Last} scores. 
Empirically, we find that the adapter outputs become more orthogonal to the prefixes when trained with RWM, indicating that the adapter complements the prefix by extending beyond the subspace spanned by a fixed number of prefix vectors.

\vspace{2pt}\noindent\textbf{Analysis of Weight Distribution.} 
Fig.~\ref{fig:weights_analysis} presents an analysis of the weights assigned to the prefixes and the adapters. We compute the average weights for each and visualize their total sums and relative ratios.
This suggests that the model can distinguish between learned and unseen tasks, highlighting the suitability of our method for CL scenarios for VLMs.
Moreover, we observe that trained tasks rely more heavily on prefix information, whereas zero-shot tasks depend primarily on the adapter.
This adaptive weighting mechanism represents a core property of our framework: when the model encounters distributions that deviate from the training data, it decreases reliance on the prefixes and activates the adapter, which is designed to preserve generalizable knowledge.

\begin{figure}[t!]
\begin{center}
\includegraphics[width=0.95\columnwidth]{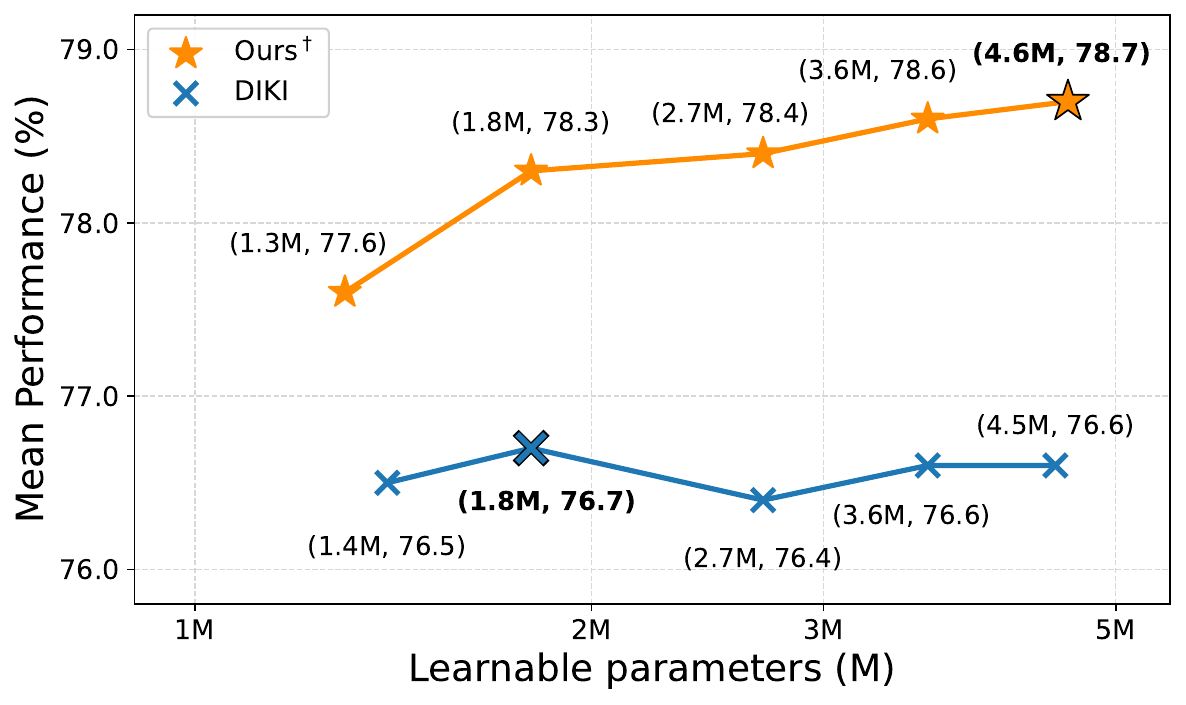}
    \vspace{-10pt}
    \caption{Comparison of mean performance and the number of learnable parameters between Ours$^{\dagger}$ and DIKI~\cite{Tang2024} under varying prefix lengths. \textbf{Bold} indicates configurations reported in Tab.~\ref{tab:mtli-order1}. Our method consistently outperforms DIKI across all settings.}
    \label{fig:num_param}
\vspace{-23.0pt}
\end{center}
\end{figure}

\vspace{2pt}\noindent\textbf{Parameter Efficiency.}
Fig.~\ref{fig:num_param} shows how performance scales with the number of learnable parameters for both our method and the baseline DIKI~\cite{Tang2024}. We control the parameter count by varying the number of prefixes $L$ in each method. The results demonstrate that our approach consistently outperforms DIKI, even when using fewer learnable parameters.

\section{Conclusion}
In this work, we propose a method for CL of VLMs that overcomes the limitations of attention-based weighting by learning task-aware weights for prefixes and adapters.
Specifically, we introduce a novel mechanism that evaluates the task relevance of each token and assigns corresponding weights accordingly.
As prefix-tuning and adapters have recently gained popularity, we believe our method can be extended to improve their effectiveness beyond the CL setting.
 
\section{Acknowledgments}
This work was partially supported by the National Research Foundation of Korea (NRF) grant funded by the Ministry of Science and ICT (MSIT) of the Korean government (RS-2024-00341749), and Institute of Information \& Communications Technology Planning \& Evaluation (IITP) grant funded by MSIT (RS-2023-00259934, RS-2025-02283048). 
{
    \small
    \bibliographystyle{ieeenat_fullname}
    \bibliography{main}
}

\clearpage
\setcounter{page}{1}
\maketitlesupplementary

\appendix
\numberwithin{table}{section}
\numberwithin{figure}{section}
\numberwithin{equation}{section}

\section{Additional Experimental Results}

\begin{figure}[t!]
\begin{center}
    \includegraphics[width=0.95\columnwidth]{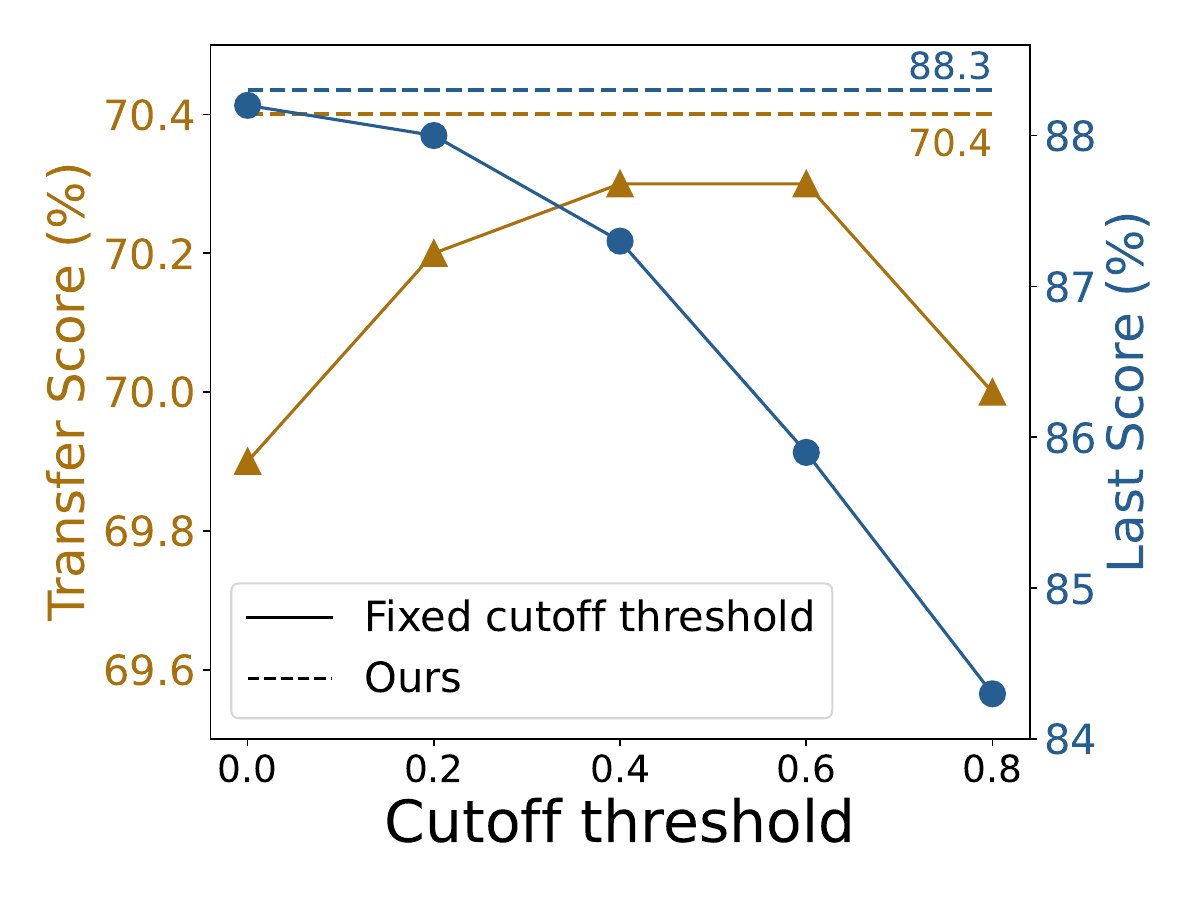}
    \vspace{-15pt}
    \caption{Comparison between fixed cutoff and the proposed dynamic cutoff strategy.}
    \label{fig:cutoff_analysis}
\vspace{-20.0pt}
\end{center}
\end{figure}

\subsection{Analysis of Conditional Filtering}

Our CondAct module refines the prefix weight \(g_{ijk}\) through a conditional filtering process, rather than directly applying the outputs of conditional normalization. This additional step selectively suppresses the influence of less relevant prefixes by zeroing out weights that fall below a certain cutoff.  
Importantly, instead of using a fixed threshold, our method determines this cutoff dynamically, based on the likelihood of each prefix score under the trained task distributions.  
In Fig.~\ref{fig:cutoff_analysis}, we compare our dynamic cutoff strategy against several fixed thresholds. In the figure, dashed lines denote the results using our dynamic approach.  
Our method consistently outperforms fixed thresholds in both \textit{Transfer} and \textit{Last} scores, clearly demonstrating the effectiveness of the proposed filtering mechanism.

\begin{figure}[t!]
\begin{center}
    \includegraphics[width=0.95\columnwidth]{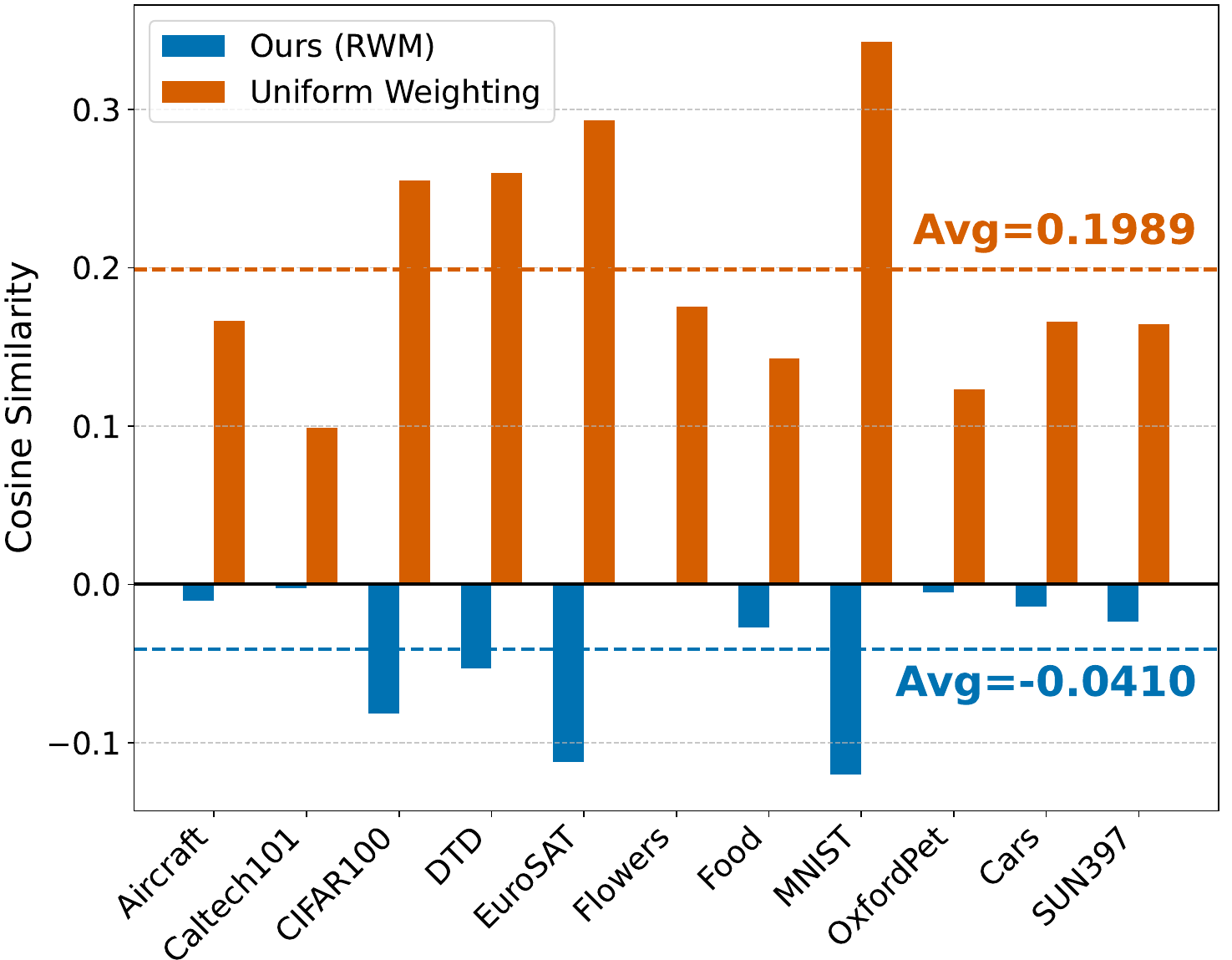}
    \vspace{-10pt}
    \caption{Cosine similarity between the prefix output and the adapter output across datasets. Our RWM yields near-orthogonal representations, whereas uniform weighting produces substantially higher cosine similarity.}
    \label{fig:orthogonality}
\vspace{-16.0pt}
\end{center}
\end{figure}

\subsection{Relationship Between Prefixes and the Adapter}

To further understand the relationship between prefixes and the adapter, we analyze the cosine similarity between \( O_{\text{prefix}_i} \), which is generated by a linear combination of prefix vectors, and the adapter output \( O_{\text{adapter}_i} \).
As shown in Fig.~\ref{fig:orthogonality}, with our residual weighting mechanism (RWM), orthogonality emerges between the prefixes and the adapter. 
This shows that the adapter complements the prefixes by generating outputs that go beyond the subspace defined by the fixed prefix vectors, thus capturing information that linear combinations of those prefixes alone cannot express.
However, as the figure shows, such orthogonality does not emerge when the adapter output is uniformly weighted to all tokens. This highlights that the proposed weighting mechanism plays a crucial role in enabling the complementary synergy between the adapter and the prefixes.

\begin{table}[t!]
\centering
\resizebox{0.9\columnwidth}{!}{%
\begin{tabular}{c|ccc}
\toprule
Weighting Method  & Transfer & Avg. & Last \\
\midrule
\( \sum_k g_{ijk} \) & 69.8 & 79.1 & 88.2 \\
\( \sum_k \widetilde{g}_{ijk} \) & 70.1 & 79.1 & 88.0 \\
\rowcolor{gray!20}
RWM (Ours) & \textbf{70.4} & \textbf{79.3} & \textbf{88.3} \\
\bottomrule
\end{tabular}
}
\vspace{-8.0pt}
\caption{Comparison of weighting strategies for adapter. \( \sum_k g_{ijk} \) and \( \sum_k \widetilde{g}_{ijk} \) use normalized and filtered prefix weights, respectively.}
\vspace{-12.0pt}
\label{tab:weighting_analysis}
\end{table}

\subsection{Comparison with Alternative Weighting Mechanisms}

Our weighting mechanism (RWM) applies token-wise weighting to the adapter's output using the total prefix score \( \sum_k \sigma(s_{ijk}) \) assigned to each input token.  
Notably, our method does not utilize the normalized attention weights \( \sum_k \widetilde{g}_{ijk} \), which are bounded within the interval \([0,1]\), but instead relies directly on the raw prefix scores prior to normalization.  
Rather than adopting an approach that directly transfers the weights assigned to the prefix onto the adapter, our method enables selective focusing by subtracting the prefix contribution from the overall activation score and using only the resulting values that remain non-negative. This design allows the adapter to selectively address tokens that are insufficiently adapted by the prefix alone.  
As illustrated in Tab.~\ref{tab:weighting_analysis}, we compare our RWM with alternative weighting strategies that employ either the sum of normalized weights \( \sum_k g_{ijk} \) or filtered weights \( \sum_k \widetilde{g}_{ijk} \), obtained through an additional filtering process. The results confirm the effectiveness of our approach, which uses the residual of the raw prefix scores.

\begin{table}[t!]
\centering
\resizebox{1.0\columnwidth}{!}{%
\begin{tabular}{l|ccc|ccc}
\toprule
Params. & Single $D$ & Frozen $D$ & SVD Init. & Transfer & Avg. & Last \\
\midrule
21.6 M & – & – & – & 69.7 & 79.0 & 88.1  \\
\midrule
11.0 M & \checkmark & – & – & 58.4 & 50.9 & 25.7 \\
11.0 M & \checkmark & \checkmark & – & 69.7 & 78.8 & 88.0 \\
\cellcolor{gray!20}11.0 M & \cellcolor{gray!20}\checkmark & \cellcolor{gray!20}\checkmark & \cellcolor{gray!20}\checkmark & \cellcolor{gray!20}\textbf{70.4} & \cellcolor{gray!20}\textbf{79.3} & \cellcolor{gray!20}\textbf{88.3} \\
\bottomrule
\end{tabular}
}
\vspace{-8.0pt}
\caption{Ablation study of adapter designs. $D$ denotes the down projection matrix. The first row uses standard LoRA, and the second row applies HydraLoRA~\cite{Tian2024HydraLoRA} with a shared, continually updated $D$ across tasks. Params.\ represents the number of additional parameters.}
\vspace{-12.0pt}
\label{tab:adapter_ablation}
\end{table}

\subsection{Analysis of Adapter Architecture}
In this paper, we adopt the HydraLoRA~\cite{Tian2024HydraLoRA} architecture as our LoRA-based adapter, which achieves parameter efficiency through a single down projection matrix \(D\). To better suit the CL scenarios, we modify this architecture by initializing \(D\) using the top-\(k\) left singular vectors of the value projection matrix from the CLIP model, and freeze this matrix throughout the CL process.
To evaluate the effectiveness of our proposed adapter architecture, we conduct an ablation study on various adapter designs, as summarized in Tab.~\ref{tab:adapter_ablation}. The first row in the table corresponds to a baseline using the standard LoRA architecture, which exhibits lower performance compared to our modified adapter. The second row shows results obtained by directly applying the original HydraLoRA~\cite{Tian2024HydraLoRA} architecture, where the \(D\) matrix is continually updated for each task. This continual update introduces catastrophic forgetting, resulting in significant performance degradation in the CL scenario.
The third row corresponds to another variant, where the down projection matrix \(D\) is randomly initialized and then frozen. In this case, the row space of \(D\) may lie within a minor subspace, and restricting fine-tuning to this subspace~\cite{Liang2024} fails to improve performance. Taken together, our proposed adapter design enables fine-tuning directly on the principal subspace of the value projection matrix, effectively leveraging the pretrained knowledge from CLIP. This approach further supports prefixes containing task-specific knowledge and facilitates additional adaptation, thereby improving performance in CL scenarios.

\begin{table}[t!]
    \centering
    \resizebox{0.7\columnwidth}{!}{
    \begin{tabular}{l|c|c}
        \toprule
        Method & \shortstack{Inference\\Time} & \shortstack{Mean\\(\%)} \\
        \midrule
        MoE-Adapters & 35m 47s & 76.9 \\
        DIKI & 3m 46s & 76.7 \\
        \cellcolor{gray!20}Ours\textdagger & \cellcolor{gray!20}\textbf{3m 8s} & \cellcolor{gray!20}78.7 \\
        \cellcolor{gray!20}Ours & \cellcolor{gray!20}3m 16s & \cellcolor{gray!20}\textbf{79.3} \\
        \bottomrule
    \end{tabular}
    }
    \vspace{-8pt}
    \caption{Comparison of methods in terms of throughput.  Mean is computed as the average of Transfer, Avg., and Last.}
    \label{tab:efficiency}
    \vspace{-8pt}
\end{table}

\begin{table}[t!]
\centering
\resizebox{0.85\columnwidth}{!}{%
\begin{tabular}{ccc|cc}
\toprule
RePA & CondAct & RWM & Inference Time & Mean (\%) \\
\midrule
- & - & - & 3m 44s & 76.9 \\
\midrule
\checkmark & - & - & 2m 56s & 77.1 \\
- & \checkmark & - & 3m 29s & 78.0 \\
\checkmark & \checkmark & - & 2m 53s & 78.9 \\
\rowcolor{gray!20}
\checkmark & \checkmark & \checkmark & 3m 8s & 79.3 \\
\bottomrule
\end{tabular}
}
\vspace{-8.0pt}
\caption{Analysis of how each component of our method contributes to the increase in inference time.}
\vspace{-10.0pt}
\label{tab:test_time_perf}
\end{table}

\subsection{Computational Costs.}
In Tab.~\ref{tab:efficiency}, we compare the inference time of our method with other PEFT approaches on the MTIL benchmark.
Our method consistently achieves both performance improvements and lower inference time.
This efficiency arises from the simplicity of our design:
(i) replacing the quadratic query–key dot-product operations with affine transformations, and
(ii) removing router operations by directly mapping the residual weights from CondAct to the adapter weights.
Together, these design choices significantly reduce computational overhead, resulting in lower inference time.

In Tab.~\ref{tab:test_time_perf}, we further analyze how each component of our method affects inference time.
Starting from the baseline, incorporating RePA substantially reduces inference time by replacing attention with lightweight affine transformations.
Using CondAct alone also provides a modest speedup by suppressing less relevant prefixes.
When combined, RePA and CondAct further reduce inference time, making the model even faster than RePA alone.
This improvement likely arises because RePA produces prefix scores that better capture task relevance, allowing CondAct to suppress more irrelevant prefixes and zero out additional prefix weights, thereby reducing effective computation.
Finally, adding RWM slightly increases inference time due to the additional adapter branch, but it achieves the best overall performance while still remaining faster than DIKI.

\section{Implementation Details}

\subsection{Experimental Setup}
We follow the baseline setup of DIKI~\cite{Tang2024} for all our experiments. Specifically, for each task, we use the same hand-crafted text prompts as DIKI, combining them with class names to construct the input to the text encoder. Following the baseline, we train the model for 10 epochs using a cosine learning rate scheduler and select the one with the highest validation accuracy, using a consistent data split throughout the process. To ensure a fair comparison, we adopt the CLIP ViT-B/16~\cite{Radford2021} model, which is commonly used by other methods on the same benchmark.

Building upon this configuration, we minimize cross-entropy loss between model predictions and ground truth labels during training. The learning rate is set to 1.25, and a batch size of 32 is used. The prefix length \(L\) is set to 8. The row rank dimension of the LoRA adapter is set to 64 in our default setting (Ours) and reduced to 4 in the parameter-efficient variant (Ours\textdagger). Both prefix and adapter modules are integrated into all 12 layers of the visual and text encoders. All experiments are conducted using a single NVIDIA 4090 GPU.
For RePA, the bias matrix \(B_i^G\) is initialized to \(-4\) for the visual branch and \(-2\) for the text branch. The weight matrix \(W_i^G\) and the prefix vector \(P_V\) are initialized as orthogonal vectors across all tasks.

\begin{table}[t!]
\centering
\resizebox{1.0\columnwidth}{!}{%
\begin{tabular}{l|ccc|c}
\toprule
Head Adjustment & Transfer (\%) & Avg (\%) & Last (\%) & Params. \\
\midrule
Original Head Count (×1)  & 69.8 & 78.9 & 88.2 & 29.6 M \\
Reduced Head Count (×0.5) & 70.1 & 79.1 & 88.1 & 21.0 M \\
Expanded Head Count (×2)  & 69.8 & 78.9 & 88.2 & 46.8 M \\
\rowcolor{gray!20}
Adaptive Head Count (Ours) & \textbf{70.4} & \textbf{79.3} & \textbf{88.3} & 30.8 M \\
\bottomrule
\end{tabular}
}
\vspace{-8.0pt}
\caption{Comparison of various head counts and our adaptive head‐count selection method.}
\label{tab:head_ablation}
\vspace{-14.0pt}
\end{table}

\subsection{Adaptive Head Count}

In this paper, we propose a novel adaptive method for determining the number of heads \( h' \) for each task, where \( h' \) is a separate set of heads used exclusively in the prefix and adapter modules, distinct from the original number of heads \( h \) in CLIP. Specifically, we compute a task-specific scaling factor that captures the discriminability of the feature representations of each task and use it to scale the original number of heads to derive \( h' \). To obtain this factor, we first measure the inter-class and intra-class variances of the features for each task.
For image features, let \(\mathbf{v}_{i,j}\) denote the normalized feature vector of the \(j\)-th sample in class \(i\), and let \(N_i\) be the number of samples in class \(i\). We compute the normalized class mean for each class \(i\) as:
\[
\hat{\mu}_i = \frac{\mu_i}{\|\mu_i\|}, \quad \text{with} \quad \mu_i = \frac{1}{N_i}\sum_{j=1}^{N_i}\mathbf{v}_{i,j}.
\]

The inter-class variance for image features is defined as one minus the average cosine similarity between different class means:
\[
\text{inter\_variance}_{\text{image}} = 1 - \frac{1}{K(K-1)} \sum_{i\neq j} \hat{\mu}_i \cdot \hat{\mu}_j,
\]
where \(K\) is the total number of classes.
The intra-class variance for image features is computed based on the average pairwise cosine similarity within each class:
{\small
\[
\text{intra\_variance}_{\text{image}} = 1 - \frac{1}{K} \sum_{i=1}^{K} \left( \frac{1}{N_i(N_i-1)} \sum_{j \neq k} \mathbf{v}_{i,j} \cdot \mathbf{v}_{i,k} \right).
\]
}

For text features, assuming that each class \(i\) is represented by a normalized embedding \(\mathbf{t}_i\), the inter-class variance is similarly defined as:
\[
\text{inter\_variance}_{\text{text}} = 1 - \frac{1}{K(K-1)} \sum_{i\neq j} \mathbf{t}_i \cdot \mathbf{t}_j.
\]

Next, we combine these variances into an \textit{easiness score}, which quantifies task difficulty by jointly considering class inter-separability and within-class consistency, with an additional correction for the number of classes. The proposed \textit{easiness score} is computed as follows:
\[
\mathrm{easiness}
= \frac{\tfrac{1}{2}\bigl(\mathrm{inter\_variance}_{\mathrm{image}}
                     + \mathrm{inter\_variance}_{\mathrm{text}}\bigr)}
       {\mathrm{intra\_variance}_{\mathrm{image}}
        + \bigl(\alpha / n_{\mathrm{cls}})},
\]
where \(n_{\mathrm{cls}}\) is the number of classes, and \(\alpha\) (set to 10) is used as a scaling factor to balance the units of the class count term.
A higher score indicates well-separated and consistent class features, implying an easier classification task.
This score is then divided by the model's zero-shot accuracy (\(\text{zs\_accuracy}\)) to yield the composite scaling factor \(F\) for each task:
\[
F = \frac{\text{easiness}}{\text{zs\_accuracy}}.
\]
Dividing the easiness score by the zero-shot performance allows the model to assess how well its pretrained knowledge aligns with each task. When a task has a high easiness score but the model shows low zero-shot accuracy, this indicates that the task is considered to require more adaptation capacity, as the pretrained features are not sufficiently informative.
The original number of heads \(h\) is scaled by \(F\) and adjusted to the nearest power of two to maintain architectural consistency:
\[
h' = 
\begin{cases}
1, & \text{if } h \times F = 0, \\
2^{\text{round}(\log_2(h \times F))}, & \text{otherwise.}
\end{cases}
\]
As shown in Tab.~\ref{tab:head_ablation}, this adaptive adjustment yields superior performance compared to simply increasing the number of parameters uniformly, highlighting the effectiveness of our adaptive approach.

\section{Details about Benchmark}
\subsection{Datasets}
Both MTIL and ODCL-CIL benchmarks were evaluated using two distinct dataset orders, followed by~\cite{Zheng2023}. MTIL is designed for the task-incremental setting, while ODCL-CIL, also referred to as MCIL, targets the class-incremental setting. In Order-I, the datasets are arranged alphabetically as follows: Aircraft, Caltech101, CIFAR100, DTD, EuroSAT, Flowers, Food, MNIST, OxfordPet, StanfordCars, and SUN397. In contrast, Order-II lists the datasets in a random sequence: StanfordCars, Food, MNIST, OxfordPet, Flowers, SUN397, Aircraft, Caltech101, DTD, EuroSAT, and CIFAR100.

\subsection{Metrics}
We further formulate the \textit{Transfer}, \textit{Avg}, and \textit{Last} metrics, which were originally introduced in~\cite{Zheng2023}. Let \(p_j^{(i)}\) denote the accuracy achieved on task \(j\) after the model has been trained on task \(i\). Given a total of \(T\) tasks, these metrics are computed as follows.

The \textbf{Transfer} metric, which measures the model's forward forgetting by evaluating its zero-shot performance after completing training on task \(j\) is defined as:
\[
\text{Transfer}_j = \frac{1}{j-1} \sum_{i=1}^{j-1} p_j^{(i)}, \quad j = 2, 3, \ldots, T.
\]
The \textbf{Avg} metric, representing the average performance across all tasks, is defined as:
\[
\text{Avg}_j = \frac{1}{T} \sum_{i=1}^{T} p_j^{(i)}, \quad j = 1, 2, \ldots, T.
\]
Finally, the \textbf{Last} metric, which reflects the final performance after training on all tasks, is given by:
\[
\text{Last}_j = p_j^{(T)}, \quad j = 1, 2, \ldots, T.
\]

\begin{table}[t!]
\centering
\resizebox{0.9\columnwidth}{!}{%
\begin{tabular}{c|ccc|c}
\toprule
Method & Trans & Avg & Last & Mean \\
\midrule
DIKI~\cite{Tang2024} & 68.7 & 72.5 & 78.6 & 73.3 \\
MoE-Adapter~\cite{Yu2024} & 69.3 & 75.3 & 82.3 & 75.6 \\
\cellcolor{gray!20}Ours & \cellcolor{gray!20}\textbf{69.6} & \cellcolor{gray!20}\textbf{75.9} & \cellcolor{gray!20}\textbf{83.7} & \cellcolor{gray!20}\textbf{76.4 \textcolor{red}{(+0.8)}} \\
\bottomrule
\end{tabular}
}
\vspace{-8.0pt}
\caption{Comparison with SOTA methods on 16-shot MTIL-FS benchmark. Results for DIKI and MoE-Adapter are based on our replications.}
\label{tab:16shots}
\vspace{-14.0pt}
\end{table}

\subsection{Few-Shot Setting}

In contrast to the baseline few-shot setup used in DIKI~\cite{Tang2024}, we conduct experiments across all datasets included in the MTIL benchmark. Specifically, DIKI excluded certain datasets, arguing that prompt-based CL methods fail to capture sufficient information from few-shot samples in some tasks. However, our method effectively addresses this limitation through a conditional filtering process that suppresses the influence of noisy prefixes while simultaneously leveraging adapters to supplement additional required information. For a fair comparison, we exclude GIFT~\cite{GIFT}, which utilizes additional synthetic data generated via diffusion models.
Instead, we replicate MoE-Adapter~\cite{Yu2024}, another PEFT-based method, alongside DIKI~\cite{Tang2024}, adapting both to our experimental setting for comparison.
A summary of the results is provided in Tab.\ref{tab:16shots}, with the complete table in Tab.\ref{tab:16shots_full}. In the few-shot setting, our method outperforms other SOTA methods, demonstrating non-trivial improvements. These results indicate that our approach performs well even in data-scarce environments.

\subsection{Detailed Results}
Tab.~\ref{tab:mtil-order2} shows the performance comparison with SOTA methods on the second-order setting of the MTIL benchmark. Our method consistently outperforms other approaches in the second-order setting as well. The complete results for both orders can be found in Tab.~\ref{tab:main_results_full} and Tab~.\ref{tab:main_results_full_order2}, respectively. The complete results for the ODCL-CIL setting are included in Tab.~\ref{tab:odcl_main_results_full}.

\begin{figure*}[t!]
\begin{center}
\resizebox{.8\linewidth}{!}{%
\includegraphics[width=0.6\textwidth]{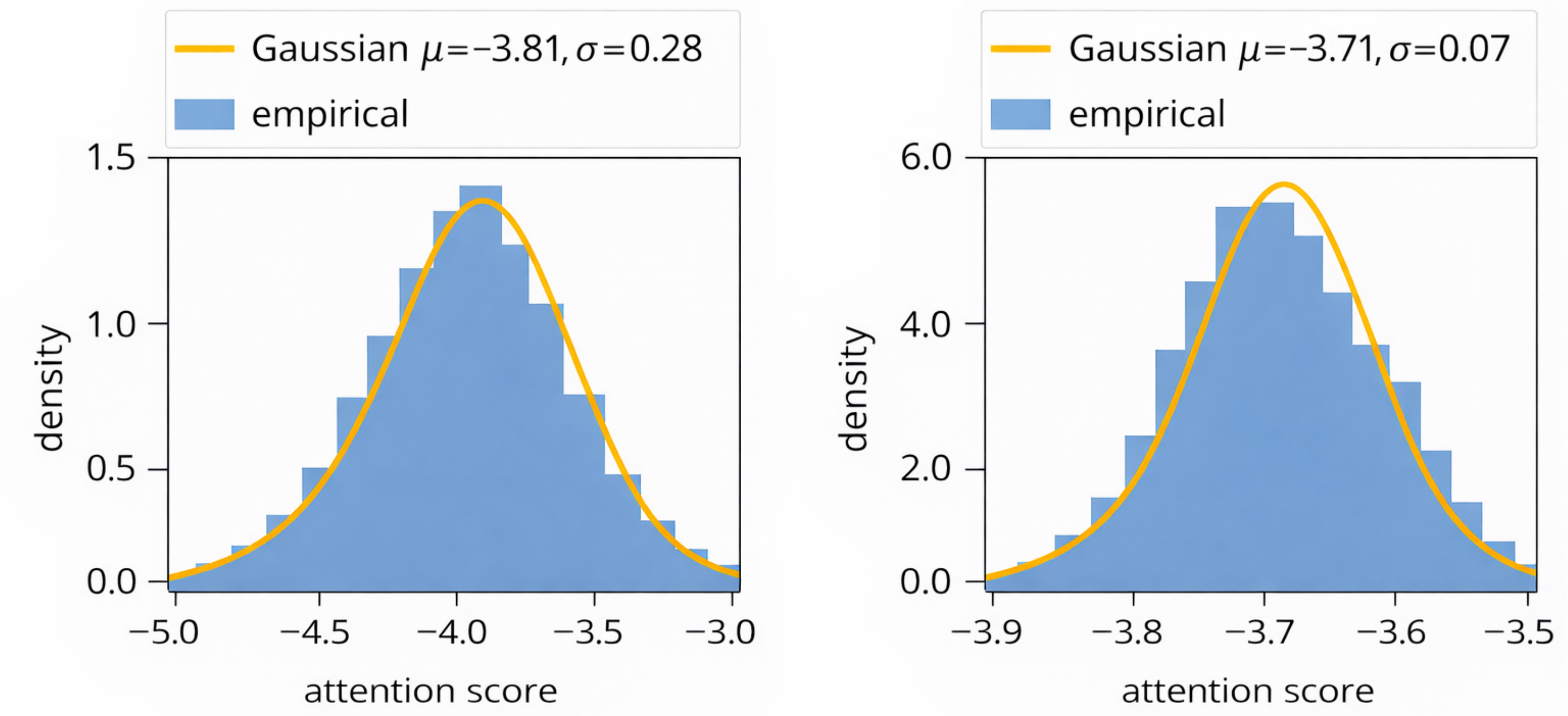}
}
\vspace{-8.0pt}
\caption{Mean and variance of attention scores between prefix tokens and the [CLS] token measured on the Aircraft dataset. The left plot shows results from Layer~3, and the right plot shows results from Layer~9. The observed distributions closely match their Gaussian counterparts, supporting the validity of the proposed cutoff design.}
\vspace{-16.0pt}
\label{fig:score_dist}
\end{center}
\end{figure*}

\section{Theoretical Analysis of CondAct for Forward Forgetting}
In this section, we provide a theoretical analysis demonstrating how our proposed CondAct mechanism mitigates forward forgetting by bounding the prefix-induced drift in token-level representations.

In the pretrained model, the token-level representation of the $j$-th input token at the $i$-th attention head is computed as
\begin{align*}
   h_{ij}^{orig} = \sum_{u=1}^m \alpha_{iju} V_{iu}, \quad \text{where} \sum_{k=1}^m 
   \alpha_{ijk} = 1.
\end{align*}
Here, $\alpha_{ijk}$ denotes the attention weights, and $V_{iu}$ denotes the corresponding projected value vectors. 
We define the prefix-induced representation as:
\begin{align*}
    p_{ij} &= \sum_{k=1}^L w_{ijk}P_{V_k}^{(i)},
\end{align*}
where $P_{V_k}^{(i)}$ is the value vector associated with the $k$-th prefix token, and $w_{ijk}$ represents the attention weight between the $j$-th input token and $k$-th prefix token.
These weights are obtained by applying a sigmoid gate to the attention score.
Specifically, let $s_{ijk}$ denote the attention score between the $j$-th input token and the $k$-th prefix token; then we consider two cases:
\begin{align*}
    w_{ijk} &= \begin{cases}
\sigma(s_{ijk}), & \text{without CondAct, Sigmoid-only.}
\\ g_{ijk} \ \text{or} \ \tilde{g}_{ijk}, & \text{with CondAct, \cref{eq:condact}}.
\end{cases}
\end{align*}

\noindent\textbf{Assumption (Bounded prefix values).}
\emph{There exists a constant $c_V > 0$ such that}
\begin{align*}
  \bigl\| P_{V_k}^{(i)} \bigr\|_2 \le c_V
  \quad \text{for all prefix tokens $k$.}
  \label{eq:appendix:bounded_prefix_values}
\end{align*}

\medskip
\noindent\textbf{Proposition (Prefix-induced drift bound).}
\emph{For any token $j$ at head $i$, the norm of the prefix-induced representation $p_{ij}$ satisfies}
\begin{align*}
  \bigl\| p_{ij} \bigr\|_2
  &\le L c_V
  \quad \text{(without CondAct, Sigmoid-only)}, \\
  \bigl\| p_{ij} \bigr\|_2
  &\le c_V
  \quad \text{(with CondAct, \cref{eq:condact})}.
\end{align*}
In particular, CondAct reduces the worst-case prefix-induced drift per token from $L c_V$ to $c_V$.

\medskip
\noindent\textit{Proof.}
By the definition of prefix-induced representation,
\begin{align*}
  \bigl\| p_{ij} \bigr\|_2
    &= \Bigl\| \sum_{k=1}^L w_{ijk} \, P_{V_k}^{(i)} \Bigr\|_2
     \le \sum_{k=1}^L |w_{ijk}| \, \bigl\| P_{V_k}^{(i)} \bigr\|_2,
\end{align*}
where we used the triangle inequality. In the Sigmoid-only setting, $w_{ijk} = \sigma(s_{ijk}) \in [0,1]$, so $|w_{ijk}| \le 1$ for all $k$. Together with the previous assumption, this yields
\begin{align*}
  \bigl\| p_{ij} \bigr\|_2
    \le \sum_{k=1}^L \bigl\| P_{V_k}^{(i)} \bigr\|_2
    \le \sum_{k=1}^L c_V
    = L c_V,
\end{align*}

We now consider the CondAct case. By construction (\cref{eq:condact:condnorm}), CondAct applies conditional normalization such that the total prefix weight per token is explicitly bounded, 
\begin{align*}
  \sum_{k=1}^L g_{ijk} \le 1,
  \qquad
  g_{ijk} \ge 0
  \quad \text{for all $i,j,k$}.
\end{align*}
In this case as well, under the same assumption as above, we obtain
\begin{align*}
  \bigl\| p_{ij} \bigr\|_2
    &\le \sum_{k=1}^L g_{ijk} \, \bigl\| P_{V_k}^{(i)} \bigr\|_2
     \le c_V \sum_{k=1}^L g_{ijk}
     \le c_V,
\end{align*}
\hfill$\square$

\medskip
\noindent\textbf{Corollary (Relative prefix-induced drift).}
\emph{Suppose the pretrained representation satisfies}
\begin{align*}
  \bigl\| h_{ij}^{\mathrm{orig}} \bigr\|_2 \ge c_h > 0
\end{align*}
\emph{for some constant $c_h$. Then the relative drift satisfies}
\begin{align*}
  \frac{\bigl\| p_{ij} \bigr\|_2}{\bigl\| h_{ij}^{\mathrm{orig}} \bigr\|_2}
  \le
  \begin{cases}
   L \cdot \dfrac{ c_V}{c_h}, & \text{without CondAct}, \\[6pt]
    \dfrac{c_V}{c_h}, & \text{with CondAct}.
  \end{cases}
\end{align*}
In summary, CondAct reduces the worst-case relative drift by a factor of up to $L$.

\section{Prefix Score Distribution}
In Eq.~\eqref{eq:cutoff_nontxt}, we leverage the empirical observation that the attention scores between prefix tokens and the [CLS] token follow a Gaussian distribution, which enables us to determine the distribution used for computing the cutoff.
To validate this assumption, we measure the prefix scores on the Aircraft dataset at Layers 3 and 9, which we selected arbitrarily.
As shown in Fig.~\ref{fig:score_dist}, the empirical distributions of the prefix scores at both layers closely align with the corresponding Gaussian fits, indicating that the Gaussian assumption provides a reasonable approximation. This observation supports the stability of our cutoff design.

\begin{table*}[t!]
    \centering
    \resizebox{\textwidth}{!}{
        \begin{tabular}{l|lccccccccccc|c}
        \toprule
        \textbf{Task} & \textbf{Method} & \textbf{Aircraft} & \textbf{Caltech101} & \textbf{CIFAR100} & \textbf{DTD} & \textbf{EuroSAT} & \textbf{Flowers} & \textbf{Food} & \textbf{MNIST} & \textbf{OxfordPet} & \textbf{Cars} & \textbf{SUN397} & \textbf{Average} \\
        \midrule
        \multirow{12}{*}{\raisebox{-7.0em}{\rotatebox{90}{\textbf{\LARGE MTIL-FS}}}} & \textbf{\large{Transfer}} \\
        & DIKI~\cite{Tang2024} & - & 92.7 & 68.6 & 44.1 & 47.8 & 69.9 & 86.1 & 58.4 & 89.0 & 65.1 & 65.2 & 68.7 \\
        & MoE-Adapters~\cite{Yu2024} & - & 88.4 & 68.2 & 42.9 & 53.5 & 70.1 & 88.5 & 63.3 & 89.0 & 64.7 & 65.1 & 69.3 \\
        & \cellcolor{gray!20}Ours & \cellcolor{gray!20}- & \cellcolor{gray!20}\textbf{92.9} & \cellcolor{gray!20}68.9 & \cellcolor{gray!20}45.2 & \cellcolor{gray!20}\textbf{53.7} & \cellcolor{gray!20}\textbf{71.1} & \cellcolor{gray!20}86.4 & \cellcolor{gray!20}60.4 & \cellcolor{gray!20}89.7 & \cellcolor{gray!20}66.0 & \cellcolor{gray!20}65.8 & \cellcolor{gray!20}\textbf{69.6 \cellcolor{gray!20}\textcolor{red}{(+0.3)}} \\
        \cmidrule(lr){2-14}
        & \textbf{\large{Avg.}} \\
        & DIKI~\cite{Tang2024} & 39.7 & 95.6 & 76.7 & 61.4 & 61.2 & 83.3 & 86.3 & 69.3 & 90.2 & 67.6 & 66.1 & 72.5 \\
        & MoE-Adapters~\cite{Yu2024} & 43.3 & 92.1 & 77.2 & 66.7 & 80.0 & 84.8 & 88.3 & 74.4 & 89.0 & 67.1 & 65.9 & 75.3  \\
        & \cellcolor{gray!20}Ours & \cellcolor{gray!20}\textbf{51.3} & \cellcolor{gray!20}\textbf{95.7} & \cellcolor{gray!20}\textbf{77.5} & \cellcolor{gray!20}\textbf{65.9} & \cellcolor{gray!20}\textbf{74.1} & \cellcolor{gray!20}\textbf{85.3} & \cellcolor{gray!20}86.7 & \cellcolor{gray!20}\textbf{71.2} & \cellcolor{gray!20}91.0 & \cellcolor{gray!20}68.8 & \cellcolor{gray!20}67.1 & \cellcolor{gray!20}\textbf{75.9 \textcolor{red}{(+0.6)}} \\
        \cmidrule(lr){2-14}
        & \textbf{\large{Last}} \\
        & DIKI~\cite{Tang2024} & 39.6 & 95.8 & 78.5 & 68.0 & 67.0 & 94.4 & 86.4 & 88.3 & 93.5 & 78.6 & 74.3 & 78.6 \\
        & MoE-Adapters~\cite{Yu2024} & 43.3 & 92.6 & 79.2 & 75.6 & 95.1 & 97.1 & 88.1 & 93.8 & 89.1 & 78.0 & 73.7 & 82.3 \\
        & \cellcolor{gray!20}Ours & \cellcolor{gray!20}\textbf{51.4} & \cellcolor{gray!20}\textbf{96.0} & \cellcolor{gray!20}\textbf{79.4} & \cellcolor{gray!20}\textbf{73.8} & \cellcolor{gray!20}\textbf{90.2} & \cellcolor{gray!20}\textbf{97.4} & \cellcolor{gray!20}87.0 & \cellcolor{gray!20}\textbf{91.8} & \cellcolor{gray!20}\textbf{93.9} & \cellcolor{gray!20}83.4 & \cellcolor{gray!20}76.6 & \cellcolor{gray!20}\textbf{83.7 \textcolor{red}{(+1.4)}} \\
        \bottomrule
    \end{tabular}
        }
\caption{Comparison with SOTA methods on MTIL-FS benchmark (Order~I) in terms of ``Transfer'', ``Average'', and ``Last'' scores (\%). Best results are highlighted in \textbf{bold}. Ours\textdagger\ indicates our reduced-parameter variant.}
    \label{tab:16shots_full}
\end{table*}

\begin{table*}[t!]
\centering
\resizebox{0.8\textwidth}{!}{%
\begin{tabular}{l c c | c c c | c}
\toprule
\textbf{Method} & \textbf{Extra data} & \textbf{Train Params.} & \textbf{Transfer} & \textbf{Avg.} & \textbf{Last} & \textbf{Mean} \\
\midrule
Zero-shot           & –            & –       & 65.4 & 65.3 & 65.3 & 65.3 \\
\midrule
LwF~\cite{Liu2020}        & $\checkmark$ & 149.6 M & 53.2 & 62.2 & 71.9 & 62.4 \\
iCaRL~\cite{Rebuffi2017}  & $\checkmark$ & 149.6 M & 50.9 & 56.9 & 71.6 & 59.8 \\
WiSE-FT~\cite{Mitchell2022} & $\checkmark$ & 149.6 M & 51.0 & 61.5 & 72.2 & 61.6 \\
ZSCL~\cite{Zheng2023}     & $\checkmark$ & 149.6 M & 64.2 & 74.5 & 83.4 & 74.0 \\
DIKI~\cite{Tang2024}      & $\times$     &   1.8 M  & 64.4 & 74.5 & 85.5 & 74.8 \\
MoE-Adapter~\cite{Yu2024} & $\times$     &  59.6 M  & 64.3 & 74.7 & 84.1 & 74.4 \\
GIFT~\cite{GIFT}          & $\checkmark$ & 149.6 M & \textbf{65.9} & 75.7 & 85.3 & 75.6 \\
\midrule
\rowcolor{gray!20}
Ours\textdagger           & $\times$     &   4.6 M  & 65.3 & 76.0 & 87.6 & 76.2 \textbf{\textcolor{red}{(+0.6)}} \\
\rowcolor{gray!20}
Ours                      & $\times$     &  30.8 M  & 65.7 & \textbf{76.4} & \textbf{88.1} & \textbf{76.7 \textcolor{red}{(+1.1)}} \\
\bottomrule
\end{tabular}
}
\caption{Comparison of SOTA methods on MTIL Order~II.}
\label{tab:mtil-order2}
\end{table*}

\begin{table*}[t!]
    \centering
    \resizebox{\textwidth}{!}{
        \begin{tabular}{l|lccccccccccc|c}
        \toprule
        \textbf{Task} & \textbf{Method} & \textbf{Aircraft} & \textbf{Caltech101} & \textbf{CIFAR100} & \textbf{DTD} & \textbf{EuroSAT} & \textbf{Flowers} & \textbf{Food} & \textbf{MNIST} & \textbf{OxfordPet} & \textbf{Cars} & \textbf{SUN397} & \textbf{Average} \\
        \midrule
        \multirow{12}{*}{\raisebox{-12.0em}{\rotatebox{90}{\textbf{\LARGE MTIL}}}} & \textbf{\large{Transfer}} \\
        & ZSCL~\cite{Zheng2023} & - & 86.0 & 67.4 & 45.4 & 50.4 & 69.1 & 87.6 & 61.8 & 86.8 & 60.1 & \textbf{66.8} & 68.1 \\
        & DIKI~\cite{Tang2024} & - & \textbf{92.9} & 69.0 & 43.2 & 48.2 & 67.4 & 85.2 & 63.0 & 87.9 & 63.8 & 66.2 & 68.7 \\
        & MoE-Adapters~\cite{Yu2024} & - & 87.9 & 68.2 & 44.4 & 49.9 & 70.7 & \textbf{88.7} & 59.7 & 89.1 & 64.5 & 65.5 & 68.9 \\
        & GIFT~\cite{GIFT} &  - & 88.5 & \textbf{69.8} & \textbf{46.0} & 49.4 & 68.5 & 87.1 & \textbf{69.9} & 88.9 & 57.7 & 67.7 & 69.3 \\
        &  \cellcolor{gray!20}Ours\textdagger &  \cellcolor{gray!20}- &  \cellcolor{gray!20}\textbf{92.9} &  \cellcolor{gray!20}68.9 &  \cellcolor{gray!20}45.2 &  \cellcolor{gray!20}53.7 &  \cellcolor{gray!20}\textbf{71.1} &  \cellcolor{gray!20}86.4 &  \cellcolor{gray!20}60.4 &  \cellcolor{gray!20}\textbf{89.7} &  \cellcolor{gray!20}\textbf{66.0} &  \cellcolor{gray!20}65.8 &  \cellcolor{gray!20}70.0 \textbf{\textcolor{red}{(+0.7)}} \\
        & \cellcolor{gray!20}Ours & \cellcolor{gray!20}- & \cellcolor{gray!20}\textbf{92.9} & \cellcolor{gray!20}69.0 & \cellcolor{gray!20}45.3 & \cellcolor{gray!20}\textbf{54.2} & \cellcolor{gray!20}\textbf{71.1} & \cellcolor{gray!20}86.2 & \cellcolor{gray!20}63.8 & \cellcolor{gray!20}89.3 & \cellcolor{gray!20}65.7 & \cellcolor{gray!20}66.5 & \cellcolor{gray!20}\textbf{70.4 \cellcolor{gray!20}\textcolor{red}{(+1.1)}} \\
        \cmidrule(lr){2-14}
        & \textbf{\large{Avg.}} \\
        & ZSCL~\cite{Zheng2023} & 45.1 & 92.0 & 80.1 & 64.3 & 79.5 & 81.6 & \textbf{89.6} & 75.2 & 88.9 & 64.7 & \textbf{68.0} & 75.4 \\
        & DIKI~\cite{Tang2024} & 45.1 & 95.5 & 83.1 & 64.8 & 79.9 & 83.5 & 87.0 & 76.2 & 89.6 & 67.0 & 67.1 & 76.3 \\
        & MoE-Adapters~\cite{Yu2024} & 50.2 & 91.9 & 83.1 & 69.4 & 78.9 & 84.0 & 89.1 & 73.7 & 89.3 & 67.7 & 66.9 & 76.7 \\
        & GIFT~\cite{GIFT} & 51.9 & 93.9 & 81.4 & 67.7 & 80.3 & 82.8 & 89.3 & 80.6 & 90.3 & 63.1 & 68.9 & 77.3 \\
        & \cellcolor{gray!20}Ours\textdagger & \cellcolor{gray!20}57.8 & \cellcolor{gray!20}\textbf{96.2} & \cellcolor{gray!20}83.9 & \cellcolor{gray!20}69.2 & \cellcolor{gray!20}82.1 & \cellcolor{gray!20}\textbf{85.8} & \cellcolor{gray!20}87.8 & \cellcolor{gray!20}74.6 & \cellcolor{gray!20}\textbf{91.2} & \cellcolor{gray!20}\textbf{69.6} & \cellcolor{gray!20}67.0 & \cellcolor{gray!20}78.6 \textbf{\textcolor{red}{(+1.3)}} \\
        & \cellcolor{gray!20}Ours & \cellcolor{gray!20}\textbf{60.4} & \cellcolor{gray!20}\textbf{96.2} & \cellcolor{gray!20}\textbf{84.5} & \cellcolor{gray!20}\textbf{70.8} & \cellcolor{gray!20}\textbf{82.4} & \cellcolor{gray!20}\textbf{85.8} & \cellcolor{gray!20}87.7 & \cellcolor{gray!20}\textbf{76.8} & \cellcolor{gray!20}90.9 & \cellcolor{gray!20}69.4 & \cellcolor{gray!20}67.6 & \cellcolor{gray!20}\textbf{79.3 \textcolor{red}{(+2.0)}} \\
        \cmidrule(lr){2-14}
        & \textbf{\large{Last}} \\
        & ZSCL~\cite{Zheng2023} & 40.6 & 92.2 & 81.3 & 70.5 & 94.8 & 90.5 & \textbf{91.9} & 98.7 & 93.9 & 85.3 & 80.2 & 83.6 \\
        & DIKI~\cite{Tang2024} & 45.2 & 95.7 & 86.3 & 72.9 & 98.0 & 97.0 & 89.2 & 99.4 & 94.2 & 81.6 & 76.6 & 85.1 \\
        & MoE-Adapters~\cite{Yu2024} & 49.8 & 92.2 & 86.1 & 78.1 & 95.7 & 94.3 & 89.5 & 98.1 & 89.9 & 81.6 & 80.0 & 85.0 \\
        & GIFT~\cite{GIFT} & 47.9 & 95.6 & 82.8 & 75.1 & 97.3 & 94.2 & 91.7 & 99.2 & 94.2 & \textbf{87.0} & \textbf{80.9} & 86.0 \\
        & \cellcolor{gray!20}Ours\textdagger & \cellcolor{gray!20}57.8 & \cellcolor{gray!20}96.5 & \cellcolor{gray!20}87.3 & \cellcolor{gray!20}78.2 & \cellcolor{gray!20}98.3 & \cellcolor{gray!20}\textbf{98.1} & \cellcolor{gray!20}89.5 & \cellcolor{gray!20}99.5 & \cellcolor{gray!20}94.9 & \cellcolor{gray!20}85.6 & \cellcolor{gray!20}78.1 & \cellcolor{gray!20}87.6 \textbf{\textcolor{red}{(+1.6)}}\\
        & \cellcolor{gray!20}Ours & \cellcolor{gray!20}\textbf{60.5} & \cellcolor{gray!20}\textbf{96.6} & \cellcolor{gray!20}\textbf{87.9} & \cellcolor{gray!20}\textbf{80.4} & \cellcolor{gray!20}\textbf{98.5} & \cellcolor{gray!20}\textbf{98.1} & \cellcolor{gray!20}89.4 & \cellcolor{gray!20}\textbf{99.6} & \cellcolor{gray!20}\textbf{95.1} & \cellcolor{gray!20}86.3 & \cellcolor{gray!20}78.8 & \cellcolor{gray!20}\textbf{88.3 \textcolor{red}{(+2.3)}} \\
        \bottomrule
    \end{tabular}
        }
\caption{Comparison with SOTA methods on MTIL benchmark (Order~I) in terms of ``Transfer'', ``Average'', and ``Last'' scores (\%). Best results are highlighted in \textbf{bold}. Ours\textdagger\ indicates our reduced-parameter variant.}
    \label{tab:main_results_full}
\end{table*}

\begin{table*}[t!]
    \centering
    \resizebox{\textwidth}{!}{
        \begin{tabular}{l|lccccccccccc|c}
        \toprule
        \textbf{Task} & \textbf{Method} & \textbf{Cars} & \textbf{Food} & \textbf{MNIST} & \textbf{OxfordPet} & \textbf{Flowers} & \textbf{SUN397} & \textbf{Aircraft} & \textbf{Caltech101} & \textbf{DTD} & \textbf{EuroSAT} & \textbf{CIFAR100} & \textbf{Average} \\
        \midrule
        \multirow{12}{*}{\raisebox{-12.0em}{\rotatebox{90}{\textbf{\LARGE MTIL}}}} & \textbf{\large{Transfer}} \\
        & ZSCL~\cite{Zheng2023} & - & 88.3 & 57.5 & 84.7 & 68.1 & 64.8 & 21.1 & 88.2 & \textbf{45.3} & \textbf{55.2} & 68.2 & 64.2 \\
        & DIKI~\cite{Tang2024} & - & 85.8 & 59.8 & 89.1 & 71.8 & 62.6 & 24.3 & 93.3 & 42.7 & 46.8 & 67.8 & 64.4 \\
        & MoE-Adapters~\cite{Yu2024} & - & \textbf{88.8} & 59.5 & 89.1 & 69.9 & 64.4 & 18.1 & 86.9 & 43.7 & 54.6 & 68.2 & 64.3 \\
        & GIFT~\cite{GIFT} & & 88.3 & \textbf{63.4} & 88.1 & 70.8 & \textbf{67.7} & 22.8 & 90.4 & \textbf{46.7} & 51.8 & 68.8 & \textbf{65.9} \\
        &  \cellcolor{gray!20}Ours\textdagger &  \cellcolor{gray!20}- &  \cellcolor{gray!20}86.9 &  \cellcolor{gray!20}61.6 &  \cellcolor{gray!20}89.3 &  \cellcolor{gray!20}71.3 &  \cellcolor{gray!20}63.2 &  \cellcolor{gray!20}24.3 &  \cellcolor{gray!20}93.3 &  \cellcolor{gray!20}44.8 &  \cellcolor{gray!20}\textbf{50.4} &  \cellcolor{gray!20}68.6 &  \cellcolor{gray!20} 65.3 \textcolor{blue}{(-0.6)} \\
        & \cellcolor{gray!20}Ours & \cellcolor{gray!20}- & \cellcolor{gray!20}{85.9} & \cellcolor{gray!20}63.3 & \cellcolor{gray!20}\textbf{89.4} & \cellcolor{gray!20}\textbf{72.2} & \cellcolor{gray!20}64.2 & \cellcolor{gray!20}\textbf{24.6} & \cellcolor{gray!20}\textbf{94.0} & \cellcolor{gray!20}44.6 & \cellcolor{gray!20}49.4 & \cellcolor{gray!20}\textbf{69.1} & \cellcolor{gray!20}65.7 \cellcolor{gray!20}\textcolor{blue}{(-0.2)} \\
        \cmidrule(lr){2-14}
        & \textbf{\large{Avg.}} \\
        & ZSCL~\cite{Zheng2023} & 81.7 & \textbf{91.3} & 91.1 & 91.0 & 82.9 & 72.5 & 33.6 & 89.7 & 53.3 & \textbf{62.8} & 69.9 & 74.5 \\
        & DIKI~\cite{Tang2024} & 81.9 & 88.9 & 92.1 & 92.8 & 87.7 & 70.3 & 34.3 & 94.2 & 51.5 & 56.1 & 69.5 & 74.5 \\
        & MoE-Adapters~\cite{Yu2024} & 84.9 & 89.9 & 89.3 & 91.4 & 86.2 & 72.2 & 33.4 & 89.4 & 53.3 & 61.4 & 69.9 & 74.7 \\
        & GIFT~\cite{GIFT} & 83.2 & 90.8 & 92.6 & 92.8 & 85.8 & \textbf{74.1} & 36.0 & 92.1 & \textbf{54.7} & 60.0 & 70.4 & 75.7 \\
        & \cellcolor{gray!20}Ours\textdagger & \cellcolor{gray!20}85.7 & \cellcolor{gray!20}89.1 & \cellcolor{gray!20}92.4 & \cellcolor{gray!20}93.2 & \cellcolor{gray!20}88.4 & \cellcolor{gray!20}71.3 & \cellcolor{gray!20}39.1 & \cellcolor{gray!20}94.5 & \cellcolor{gray!20}53.2 & \cellcolor{gray!20}59.1 & \cellcolor{gray!20}70.3 & \cellcolor{gray!20}76.0 \textbf{\textcolor{red}{(+0.3)}} \\
        & \cellcolor{gray!20}Ours & \cellcolor{gray!20}\textbf{86.6} & \cellcolor{gray!20}88.9 & \cellcolor{gray!20}\textbf{92.8} & \cellcolor{gray!20}\textbf{93.6} & \cellcolor{gray!20}\textbf{88.5} & \cellcolor{gray!20}71.5 & \cellcolor{gray!20}\textbf{40.4} & \cellcolor{gray!20}\textbf{94.9} & \cellcolor{gray!20}54.2 & \cellcolor{gray!20}58.3 & \cellcolor{gray!20}\textbf{70.9} & \cellcolor{gray!20}\textbf{76.4 \textcolor{red}{(+0.7)}} \\
        \cmidrule(lr){2-14}
        & \textbf{\large{Last}} \\
        & ZSCL~\cite{Zheng2023} & 78.2 & \textbf{91.1} & 97.6 & 92.5 & 87.4 & 78.2 & 45.0 & 92.3 & 72.7 & 96.2 & 86.3 & 83.4 \\
        & DIKI~\cite{Tang2024} & 81.9 & 89.2 & 99.4 & 94.3 & 96.8 & 76.7 & 46.3 & 95.9 & 74.8 & 98.3 & 86.6 & 85.5 \\
        & MoE-Adapters~\cite{Yu2024} & 84.1 & 88.5 & 94.0 & 91.8 & 94.1 & 77.8 & 50.4 & 93.3 & 77.1 & 87.7 & 86.6 & 84.1 \\
        & GIFT~\cite{GIFT} & 81.0 & 90.2 & 98.6 & 94.0 & 91.5 & \textbf{78.6} & 51.7 & 94.6 & 75.6 & 95.4 & 86.6 & 85.3 \\
        & \cellcolor{gray!20}Ours\textdagger & \cellcolor{gray!20}85.7 & \cellcolor{gray!20}89.4 & \cellcolor{gray!20}99.5 & \cellcolor{gray!20}94.7 & \cellcolor{gray!20}\textbf{98.1} & \cellcolor{gray!20}78.1 & \cellcolor{gray!20}56.9 & \cellcolor{gray!20}\textbf{96.6} & \cellcolor{gray!20}78.4 & \cellcolor{gray!20}\textbf{98.4} & \cellcolor{gray!20}87.3 & \cellcolor{gray!20}87.6 \cellcolor{gray!20}\textbf{\textcolor{red}{(+2.3 )}} \\
        & \cellcolor{gray!20}Ours & \cellcolor{gray!20}\textbf{86.6} & \cellcolor{gray!20}89.2 & \cellcolor{gray!20}\textbf{99.6} & \cellcolor{gray!20}\textbf{95.1} & \cellcolor{gray!20}97.9 & \cellcolor{gray!20}78.4 & \cellcolor{gray!20}\textbf{59.4} & \cellcolor{gray!20}96.4 & \cellcolor{gray!20}\textbf{79.8} & \cellcolor{gray!20}\textbf{98.4} & \cellcolor{gray!20}\textbf{87.9} & \cellcolor{gray!20}\textbf{88.1 \textcolor{red}{(+2.8)}} \\
        \bottomrule
    \end{tabular}
        }
\caption{Comparison with SOTA methods on MTIL benchmark (Order~II) in terms of ``Transfer'', ``Average'', and ``Last'' scores (\%). Best results are highlighted in \textbf{bold}. Ours\textdagger\ indicates our reduced-parameter variant.}
    \label{tab:main_results_full_order2}
\end{table*}

\begin{table*}[t!]
    \centering
    \resizebox{\textwidth}{!}{
        \begin{tabular}{l|lccccccccccc|c}
        \toprule
        \textbf{Task} & \textbf{Method} & \textbf{Aircraft} & \textbf{Caltech101} & \textbf{CIFAR100} & \textbf{DTD} & \textbf{EuroSAT} & \textbf{Flowers} & \textbf{Food} & \textbf{MNIST} & \textbf{OxfordPet} & \textbf{Cars} & \textbf{SUN397} & \textbf{Average} \\
        \midrule
        \multirow{12}{*}{\raisebox{-18.0em}{\rotatebox{90}{\textbf{\LARGE ODCL-CIL (MCIL)}}}} & \textbf{\large{Transfer}} \\
        & ZSCL~\cite{Zheng2023} & - & 84.6 & 67.5 & 44.8 & 51.5 & 69.0 & 87.6 & 62.3 & 87.1 & 59.7 & 66.4 & 68.0 \\
        & MoE-Adapters~\cite{Yu2024} & - & 88.2 & 66.8 & 44.7 & 54.1 & 70.6 & 88.4 & 59.5 & 89.0 & 64.7 & 65.0 & 69.1 \\
        & CoLeCLIP~\cite{Li2024} & - & 88.2 & 65.1 & 44.7 & 54.1 & 68.8 & \textbf{88.5} & 59.5 & 89.0 & 64.7 & 65.1 & 68.8 \\
        & DPeCLIP~\cite{Lu2024} & - & 88.2 & 67.2 & 44.7 & 54.0 & 70.6 & 88.2 & 59.5 & 89.0 & 64.7 & 64.8 & 69.1 \\
        &  \cellcolor{gray!20}Ours\textdagger &  \cellcolor{gray!20}- &  \cellcolor{gray!20}\textbf{92.9} &  \cellcolor{gray!20}68.9 &  \cellcolor{gray!20}45.2 &  \cellcolor{gray!20}53.7 &  \cellcolor{gray!20}\textbf{71.1} &  \cellcolor{gray!20}86.4 &  \cellcolor{gray!20}60.4 &  \cellcolor{gray!20}\textbf{89.7} &  \cellcolor{gray!20}\textbf{66.0} &  \cellcolor{gray!20}65.8 &  \cellcolor{gray!20}70.0 \textbf{\textcolor{red}{(+0.9)}} \\
        & \cellcolor{gray!20}Ours & \cellcolor{gray!20}- & \cellcolor{gray!20}\textbf{92.9} & \cellcolor{gray!20}\textbf{69.0} & \cellcolor{gray!20}\textbf{45.3} & \cellcolor{gray!20}\textbf{54.2} & \cellcolor{gray!20}\textbf{71.1} & \cellcolor{gray!20}86.2 & \cellcolor{gray!20}\textbf{63.8} & \cellcolor{gray!20}89.3 & \cellcolor{gray!20}65.7 & \cellcolor{gray!20}\textbf{66.5} & \cellcolor{gray!20}\textbf{70.4 \cellcolor{gray!20}\textcolor{red}{(+1.3)}} \\
        \cmidrule(lr){2-14}
        & \textbf{\large{Avg.}} \\
        & ZSCL~\cite{Zheng2023} & 46.3 & 68.3 & 74.3 & 56.3 & 79.1 & 81.4 & 89.5 & 74.0 & 89.0 & 64.4 & 67.5 & 71.8 \\
        & MoE-Adapters~\cite{Yu2024} & 37.2 & 65.3 & 79.5 & 67.6 & 19.7 & 83.1 & 80.5 & 74.0 & 88.5 & 67.5 & 65.3 & 66.2 \\
        & CoLeCLIP~\cite{Li2024} & 48.2 & 77.8 & 71.7 & 65.7 & 76.8 & 83.8 & 89.6 & 72.2 & 90.3 & 68.0 & 66.4 & 73.7 \\
        & DPeCLIP~\cite{Lu2024} & 49.9 & 85.3 & 81.5 & 65.3 & 81.6 & 84.3 & \textbf{89.9} & 74.0 & 90.4 & 68.3 & 66.2 &  76.1 \\
        & \cellcolor{gray!20}Ours\textdagger & \cellcolor{gray!20}57.3 & \cellcolor{gray!20}92.8 & \cellcolor{gray!20}83.2 & \cellcolor{gray!20}67.5 & \cellcolor{gray!20}81.7 & \cellcolor{gray!20}\textbf{85.0} & \cellcolor{gray!20}87.7 & \cellcolor{gray!20}74.5 & \cellcolor{gray!20}\textbf{90.6} & \cellcolor{gray!20}\textbf{69.5} & \cellcolor{gray!20}66.9 & \cellcolor{gray!20}77.9 \textbf{\textcolor{red}{(+1.8)}}\\
        & \cellcolor{gray!20}Ours & \cellcolor{gray!20}\textbf{60.2} & \cellcolor{gray!20}\textbf{93.5} & \cellcolor{gray!20}\textbf{83.9} & \cellcolor{gray!20}\textbf{68.9} & \cellcolor{gray!20}\textbf{82.0} & \cellcolor{gray!20}84.8 & \cellcolor{gray!20}87.6 & \cellcolor{gray!20}\textbf{76.7} & \cellcolor{gray!20}90.4 & \cellcolor{gray!20}69.4 & \cellcolor{gray!20}\textbf{67.6} & \cellcolor{gray!20}\textbf{78.6 \cellcolor{gray!20}\textcolor{red}{(+2.5)}} \\
        \cmidrule(lr){2-14}
        & \textbf{\large{Last}} \\
        & ZSCL~\cite{Zheng2023} & 42.5 & 64.4 & 67.2 & 54.8 & 89.7 & 90.4 & 91.7 & 95.8 & 93.4 & 85.2 & 78.3 & 77.6  \\
        & MoE-Adapters~\cite{Yu2024} & 34.1 & 47.6 & 80.9 & 75.5 & 0.00 & 93.0 & 70.8 & \textbf{99.4} & 86.4 & 79.8 & 68.9 & 66.9 \\
        & CoLeCLIP~\cite{Li2024} & 48.1 & 73.1 & 65.2 & 69.6 & 84.0 & 96.2 & 90.9 & 94.6 & 93.5 & 82.6 & 79.3 & 79.7 \\
        & DPeCLIP~\cite{Lu2024} & 49.9 & 84.2 & 83.2 & 71.1 & 97.0 & 95.8 & \textbf{92.0} & \textbf{99.4} & \textbf{93.9} & 84.5 & \textbf{80.2} & 84.6 \\
        & \cellcolor{gray!20}Ours\textdagger & \cellcolor{gray!20}57.2 & \cellcolor{gray!20}89.9 & \cellcolor{gray!20}85.2 & \cellcolor{gray!20}74.1 & \cellcolor{gray!20}97.7 & \cellcolor{gray!20}\textbf{96.5} & \cellcolor{gray!20}89.1 & \cellcolor{gray!20}99.2 & \cellcolor{gray!20}92.7 & \cellcolor{gray!20}85.1 & \cellcolor{gray!20}77.4 & \cellcolor{gray!20}85.8 \textbf{\textcolor{red}{(+1.2)}}\\
        & \cellcolor{gray!20}Ours & \cellcolor{gray!20}\textbf{60.1} & \cellcolor{gray!20}\textbf{91.1} & \cellcolor{gray!20}\textbf{85.9} & \cellcolor{gray!20}\textbf{75.7} & \cellcolor{gray!20}\textbf{97.9} & \cellcolor{gray!20}96.2 & \cellcolor{gray!20}89.1 & \cellcolor{gray!20}99.2 & \cellcolor{gray!20}93.0 & \cellcolor{gray!20}\textbf{86.0} & \cellcolor{gray!20}78.3 & \cellcolor{gray!20}\textbf{86.6 \cellcolor{gray!20}\textcolor{red}{(+2.0)}} \\
        \bottomrule
    \end{tabular}
        }
\caption{Comparison with SOTA methods on ODCL-CIL benchmarks in terms of ``Transfer,'' ``Average,'' and ``Last'' scores (\%). Best results are highlighted in \textbf{bold}. Ours\textdagger\ indicates our reduced-parameter variant.}
    \label{tab:odcl_main_results_full}
\end{table*}

\end{document}